\pdfoutput=1
\documentclass[10pt,twocolumn,letterpaper]{article}

\usepackage{cvpr}      

\usepackage{threeparttable}
\usepackage{graphicx}
\usepackage{amsmath}
\usepackage{amssymb}
\usepackage{booktabs}
\usepackage{graphicx, amsmath, amssymb, caption, subcaption, multirow, overpic, textpos, multibib}
\usepackage[table]{xcolor}
\usepackage{makecell}
\usepackage{graphicx}

\makeatletter
\@namedef{ver@everyshi.sty}{}
\makeatother
\usepackage{tikz}
\usepackage{comment}
\usepackage{amsmath,amssymb} %
\usepackage{color}
\usepackage{enumitem} 
\usepackage{cuted}

\newcommand{\para}[1]{\noindent\textbf{#1}.}

\usepackage[accsupp]{axessibility}  %

\usepackage[british,english,american]{babel}
\definecolor{citecolor}{HTML}{0071BC}
\definecolor{linkcolor}{HTML}{ED1C24}
\newcommand{\app}{\raise.17ex\hbox{$\scriptstyle\sim$}}

\usepackage{pifont}
\usepackage{xspace}
\makeatletter
\DeclareRobustCommand\onedot{\futurelet\@let@token\@onedot}
\def\@onedot{\ifx\@let@token.\else.\null\fi\xspace}

\def\eg{\emph{e.g}\onedot} 
\def\ie{\emph{i.e}\onedot}

\makeatother

\newcolumntype{x}[1]{>{\centering\arraybackslash}p{#1pt}}
\newcolumntype{y}[1]{>{\raggedright\arraybackslash}p{#1pt}}
\newcolumntype{z}[1]{>{\raggedleft\arraybackslash}p{#1pt}}
\newlength\savewidth\newcommand\shline{\noalign{\global\savewidth\arrayrulewidth
  \global\arrayrulewidth 1pt}\hline\noalign{\global\arrayrulewidth\savewidth}}

\makeatletter\renewcommand\paragraph{\@startsection{paragraph}{4}{\z@}
  {.5em \@plus1ex \@minus.2ex}{-.5em}{\normalfont\normalsize\bfseries}}\makeatother

\definecolor{baselinecolor}{gray}{.9}
\definecolor{deemph}{gray}{0.6}
\newcommand{\gc}[1]{\textcolor{deemph}{#1}}

\definecolor{yellow}{rgb}{1,1, 0.7}
\definecolor{lightyellow}{rgb}{1,1, 0.8}
\definecolor{orange}{rgb}{1, 0.9, 0.82}
\definecolor{tablered}{rgb}{0.99, 0.82, 0.82}

\usepackage[pagebackref,breaklinks,colorlinks]{hyperref}
\usepackage[capitalize]{cleveref}

\crefname{section}{Sec.}{Secs.}
\Crefname{section}{Section}{Sections}
\Crefname{table}{Table}{Tables}
\crefname{table}{Tab.}{Tabs.}

\def\cvprPaperID{4093} 
\def\confName{CVPR}
\def\confYear{2023}

\begin{document}

\title{NeuFace: Realistic 3D Neural Face Rendering from Multi-view Images}

\author{
Mingwu Zheng$^{1,2}$,
Haiyu Zhang$^{1,2}$,
Hongyu Yang$^{3}$,
Di Huang$^{1,2,4}$\thanks{Corresponding author.}\\
$^{1}$State Key Laboratory of Software Development Environment, Beihang University, Beijing, China\\
$^{2}$School of Computer Science and Engineering, Beihang University, Beijing, China\\
$^{3}$Institute of Artificial Intelligence, Beihang University, Beijing, China\\
$^{4}$Hangzhou Innovation Institute, Beihang University, Hangzhou, China\\
{\tt\small \{zhengmingwu, zhyzhy, hongyuyang, dhuang\}@buaa.edu.cn}
}

\maketitle

\begin{abstract}
   Realistic face rendering from multi-view images is beneficial to various computer vision and graphics applications. Due to the complex spatially-varying reflectance properties and geometry characteristics of faces, however, it remains challenging to recover 3D facial representations both faithfully and efficiently in the current studies. This paper presents a novel 3D face rendering model, namely \textbf{\textit{NeuFace}}, to learn accurate and physically-meaningful underlying 3D representations by neural rendering techniques. It naturally incorporates the neural BRDFs into physically based rendering, capturing sophisticated facial geometry and appearance clues in a collaborative manner. Specifically, we introduce an approximated BRDF integration and a simple yet new low-rank prior, which effectively lower the ambiguities and boost the performance of the facial BRDFs. Extensive experiments demonstrate the superiority of NeuFace in human face rendering, along with a decent generalization ability to common objects. Code is released at \href{https://github.com/aejion/NeuFace}{NeuFace}.
\end{abstract}

\section{Introduction}
\label{sec:intro}
Rendering realistic human faces with controllable viewpoints and lighting is now becoming ever increasingly important with its applications ranging from game production, movie industry, to immersive experiences in the Metaverse. 
Various factors, including the sophisticated geometrical differences among individuals, the person-specific appearance idiosyncrasies, along with the spatially-varying reflectance properties of skins, collectively make faithful face rendering a rather difficult problem.

According to photogrammetry, the pioneering studies on this issue generally leverage complex active lighting setups, \eg, LightStage~\cite{debevec2000acquiring}, to build 3D face models from multiple photos of an individual, where accurate shape attributes and high-quality diffuse and specular reflectance properties are commonly acknowledged as the premises of its success. An elaborately designed workflow is required, typically involving a series of stages such as camera calibration, dynamic data acquisition, multi-view stereo, material estimation, and texture parameterization~\cite{nvidia2021realistic}. While a compelling and convincing 3D face model can be finally obtained, this output highly depends on the expertise of the engineers and artists with significant manual efforts, as the multi-step process inevitably brings diverse optimization goals.

\begin{figure}
  \centering
  \setlength{\abovecaptionskip}{0pt}
  \setlength{\belowcaptionskip}{0pt}
  \includegraphics[width=1.\linewidth]{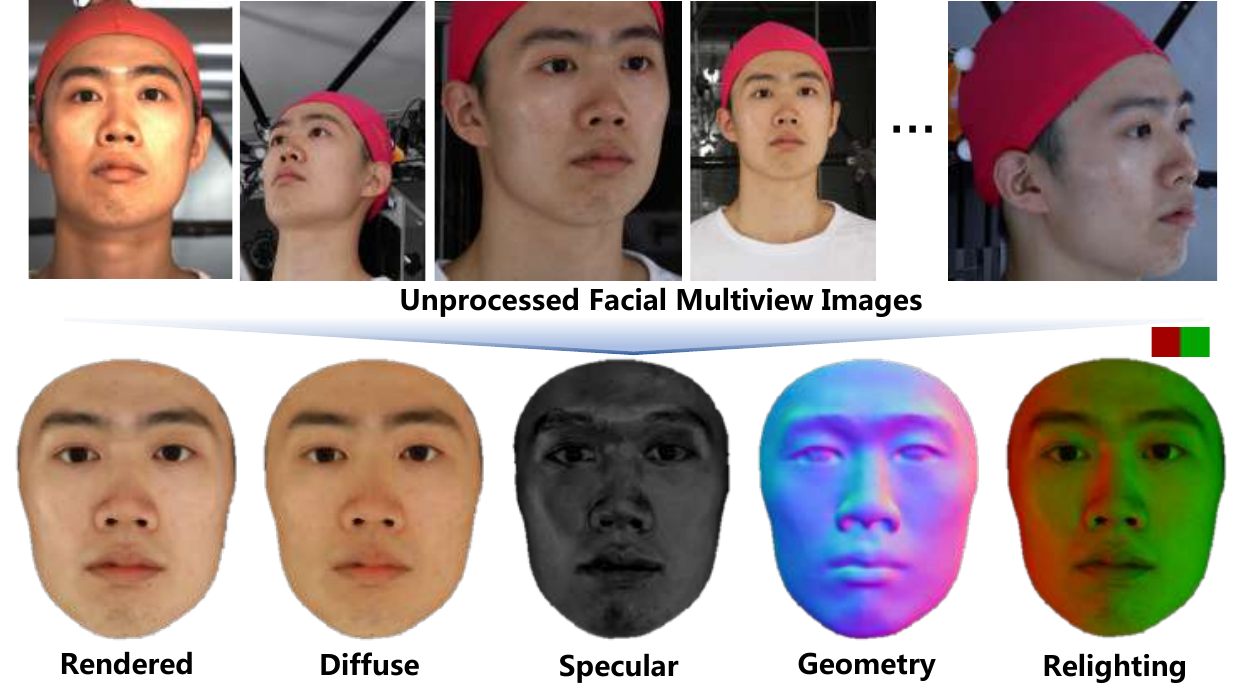}
   \caption{Demonstration of the face rendering results and recovered underlying 3D representations.}
   \label{fig:teaser}
   \vspace{-2mm}
\end{figure}

Recently, 3D neural rendering, which offers an end-to-end alternative, has demonstrated promising performance in recovering scene properties from real-world imageries, such as view-dependent radiance~\cite{verbin2022ref,martin2021nerfw,mildenhall2022nerfd,pumarola2021dnerf,reiser2021kilonerf} and geometry~\cite{yariv2020idr,yariv2021volsdf,wang2021neus,wang2022hfneus,oechsle2021unisurf}. It is mainly credited to the disentanglement of the learnable 3D representations and the differentiable image formation process, free of the tedious photogrammetry pipeline. However, like classical function fitting, inverse rendering is fundamentally under-constrained, which may incur badly-conditioned fits of the underlying 3D representations, especially for intricate cases, \eg, non-Lambertian surfaces with view-dependent highlights. With the trend in the combination of computer graphics and learning techniques, several attempts take advantage of physically motivated inductive biases and present Physically Based Rendering (PBR)~\cite{hasselgren2022shape,zhang2021physg,munkberg2022extracting,verbin2022ref}, where Bidirectional Reflectance Distribution Functions (BRDFs) are widely adopted. By explicitly mimicking the interaction of the environment light with the scene, they facilitate network optimization and deliver substantial gains. Unfortunately, the exploited physical priors are either heuristic or analytic\cite{karis2013real,torrance1967theory,cook1982reflectance}, limited to a small set of real-world materials, \eg, metal, incapable of describing human faces.

For realistic face rendering, the most fundamental issue lies in accurately modeling the optical properties of multi-layered facial skin~\cite{klehm2015recent}. In particular, the unevenly distributed fine-scale oily layers and epidermis reflect the incident lights irregularly, leading to complex view-dependent and spatially-varying highlights. This characteristic and the low-textured nature of facial surfaces strongly amplify the shape-appearance ambiguity. Moreover, subsurface scattering between the underlying dermis and other skin layers further complicates this problem. 

In this paper, we follow the PBR paradigm for its potential in learning 3D representations and make the first step towards realistic 3D neural face rendering, mainly targeting complex skin reflection modeling. Our method, namely \textit{NeuFace}, is able to recover faithful facial reflectance and geometry from only multi-view images. Concretely, we establish a PBR framework to learn neural BRDFs to describe facial skin, which simulates physically-correct light transport with a much higher representation capability. By using a differentiable Signed Distance Function (SDF) based representation, \ie \textit{ImFace}~\cite{zheng2022imface}, as the shape prior, the facial appearance and geometry field can be synchronously optimized in inverse rendering.

Compared to the analytic BRDFs, the neural ones allow richer representations for sophisticated material like facial skin. In spite of this superiority, such representations pose challenges to computational cost and data demand during training. To tackle these difficulties, the techniques in real-time rendering~\cite{akenine2019real} are adapted to separate the hemisphere integral of neural BRDFs, where the material and light integrals are individually learned instead, bypassing the massive Monte-Carlo sampling phase~\cite{pharr2016physically} required by numerical solutions. Furthermore, a low-rank prior is introduced into the spatially-varying facial BRDFs, which greatly restricts the solution space thereby diminishing the need for large-scale training observations. These model designs indeed enable \textit{NeuFace} to accurately and stably describe how the light interacts with the facial surface as in the real 3D space. Fig.~\ref{fig:teaser} displays an example.

The main contributions of this study include: 1) A novel framework with naturally-bonded PBR as well as neural BRDF representations, which collaboratively captures facial geometry and appearance properties in complicated facial skin. 2) A new and simple low-rank prior, which significantly facilitates the learning of neural BRDFs and improves the appearance recovering performance. 3) Impressive face rendering results from only multi-view images, applicable to various applications such as relighting, along with a decent generalization ability to common objects.

\section{Related Work}

We restrict the discussion specifically to static facial geometry and appearance capturing and 3D neural rendering. Please refer to \cite{klehm2015recent,egger20203dmm,tewari2022advances} for more in-depth discussion.

\textbf{Face Capturing.} Its goal is to render a realistic 3D face under arbitrary lighting condition. Existing methods generally take advantage of photogrammetry techniques to estimate facial geometries and appearances, requiring massive manual efforts. In this case, they typically decompose the problem where facial geometry is pre-captured by an intricate multi-view stereo process~\cite{beeler2010high, bradley2010high}. However, facial appearances with reflectances are still hard to acquire due to complex interactions between light and skin. The initial attempts~\cite{debevec2000acquiring, weyrich2006analysis, jensen2001practical} tackle the challenge by densely capturing the per-pixel facial reflectance at the cost of extensive data acquisition and specialized equipment. Subsequently, gradient~\cite{ma2007rapid, ghosh2011multiview, kampouris2018diffuse} or polarized illumination~\cite{ma2007rapid, ghosh2008practical, ghosh2011multiview, riviere2020single} is explored to reduce the memory cost, where most efforts are paid to the well-conditioned fitting of BRDFs. In contrast to the studies above, our solution is truly end-to-end and only observes facial skin under a single, unknown illumination condition, without cumbersome capturing settings.

\begin{figure*}
  \centering
  \includegraphics[width=1\linewidth]{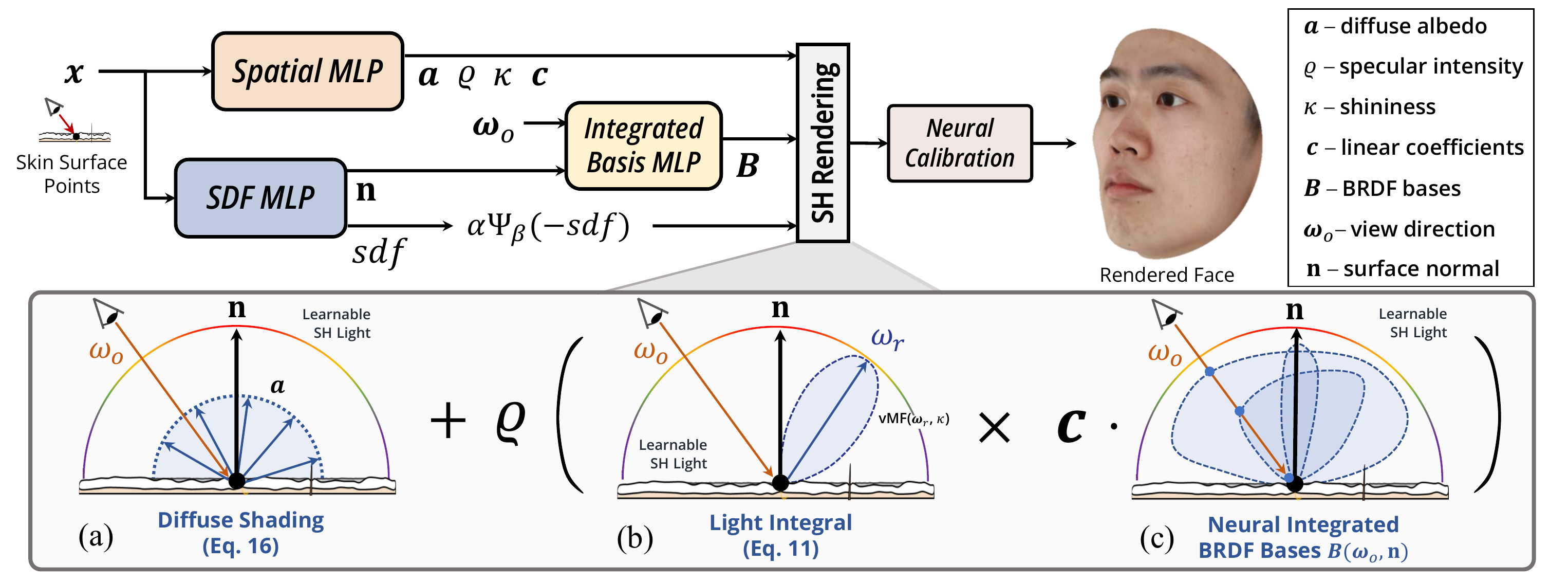}
  \caption{{\bf Method overview.} \textit{NeuFace} is composed of an appearance module implemented by \textit{Spatial MLP} and \textit{Integrated Basis MLP}, a geometry module achieved by \textit{SDF MLP}, a physically-based forward rendering, and a neural photometric calibration module. In forward rendering, the complex appearance reflectance properties are explicitly decomposed to (a) diffuse albedo, (b) light integral, and (c) BRDFs integral. Closed-form solutions are provided for the first two terms under learnable Spherical Harmonics (SH) lighting. For the BRDFs integral term, a low-rank prior is introduced to learn a set of neural bases which describe the complex material characteristics of facial skin. 
  }
  \label{fig:pipeline}
\end{figure*}

\textbf{3D Neural Rendering.} The recent advance in this field, like NeRF~\cite{mildenhall2020nerf}, have already revolutionized the paradigm of multi-view reconstruction. With learnable volumetric representations (\eg, neural field~\cite{mildenhall2020nerf, yariv2020idr}, grid~\cite{fridovich2022plenoxels}, and hybrid~\cite{mueller2022instant}) and analytic differentiable forward mapping, scene properties can be directly inferred from only 2D imageries. While reporting impressive results on novel view synthesis~\cite{verbin2022ref, martin2021nerfw, mildenhall2022nerfd, pumarola2021dnerf, reiser2021kilonerf} or geometry reconstruction~\cite{yariv2020idr, yariv2021volsdf, wang2021neus, wang2022hfneus, oechsle2021unisurf}, studies still suffer from 2D-3D ambiguities, leading to that realistic appearances and accurate geometries can hardly be established simultaneously~\cite{zhang2020nerfpp}. Such an issue is more prominent for facial skin due to the rather complicated reflection properties. Ref-NeRF~\cite{verbin2022ref} makes a step forward to the goal of both accurate surface normals and glossy appearances, which is achieved by reparameterizing the radiance in classical NeRF with reflection direction. It validates the significance of physical laws in disambiguity. \cite{hasselgren2022shape, zhang2021physg, munkberg2022extracting, sun2021nelf, li2022eyenerf} go further with PBR pipelines, delivering improved quality and supporting relighting simultaneously. Nevertheless, simplified material models are utilized or assumed, which are incapable of handling complex materials like skin. \cite{zheng2021compact, zhang2021nerfactor, lyu2022neural} train a neural material model from measurement data. But acquiring such data is impractical for live biotissue. By contrast, our method directly builds neural skin BRDFs and reconstructs accurate geometrical clues without any external data.

\section{Preliminaries}

We resort to PBR~\cite{kajiya1986rendering} to explicitly predict complex skin reflectances, where outgoing radiance $L_o$ at surface position $x$ along direction $\omega_o$ can be expressed as:
{\small
\begin{equation}
    L_o(x, \omega_o)=\int_{S^2} L_i(\omega_i) f(x, \omega_i, \omega_o) (\omega_i \cdot \mathbf{n})^{+} \mathrm{d} \omega_i ,
    \label{rendering_eq}
\end{equation}
}
\hspace{-0.5em} where $L_o$ is calculated by an integral of the product of incident radiance $L_i$ from direction $\omega_i$, skin BRDF $f$, and half-cosine function $(\omega_i \cdot \mathbf{n})^{+}$ between surface normal $\mathbf{n}$ and $\omega_i$ over sphere $S^2$. Single-bounce direct illumination without shadows is assumed, and $L_i$ is thus independent to $x$. Besides, as a pixel-level subsurface scattering is assumed, $L_o$ is only considered within a single, small region.

Spherical Harmonics (SH)~\cite{chen2019learning} and Spherical Gaussians (SG)~\cite{zhang2021physg} are commonly exploited as efficient representations of incident illumination. SH is selected in this study for its completeness and compactness. More importantly, it can implicitly facilitate specular separation with lower orders~\cite{tunwattanapong2013acquiring}. Accordingly, unknown $L_i$ can be approximated by spherical basis functions ${\mit Y}_{\ell m}$ of the first ten orders multiplied by the corresponding learnable coefficients $c_{\ell m}$:

\vspace{-2mm}
{\small
\begin{equation}
    L_i(\omega_i) \approx \sum_{\ell=0}^{\text {10}} \sum_{m=-\ell}^\ell c_{\ell m}{\mit Y}_{\ell m}(\omega_i) ,
    \label{sh_l}
\end{equation}
}
\hspace{-0.5em} $f$ is modeled as two components, akin to prior art~\cite{riviere2020single, gotardo2018practical}: 
\vspace{-1mm}
{\small
\begin{equation}
    f\left(x, \omega_i, \omega_o\right) = \frac{\mathbf{a}(x)}{\pi}+\varrho f_{\mathrm{s}}\left(x, \omega_i, \omega_o\right) ,
\end{equation}}
\hspace{-0.39em}where the left term is a diffuse component (Lambertian) that is only determined by albedo $\mathbf{a}(x)$, with the residual term mainly accounting for the glossy reflection, which is represented by $f_{\mathrm{s}}\left(x, \omega_i, \omega_o\right)$ with a scale factor $\varrho$ indicating the specular intensity. Consequently, the rendering equation (Eq.~\eqref{rendering_eq}) can be split into two terms:
\vspace{-1mm}
{\small
\begin{equation}
    \begin{split}
    & L_o(x, \omega_o)=
    \underbrace{
    \frac{{\bf a}(x)}{\pi} \int_{S^2} L_i(\omega_i) (\omega_i \cdot \mathbf{n})^{+} \mathrm{d} \omega_i
    }_{\text{diffuse term}~L_{\mathrm{d}}}\\
    & +\underbrace{
    \varrho \int_{S^2} L_i(\omega_i) f_{\mathrm{s}}(x, \omega_i, \omega_o) (\omega_i \cdot \mathbf{n})^{+} \mathrm{d} \omega_i
    }_{\text{residual (specular) term}~L_{\mathrm{s}}} .
    \label{rendering_eq2}
    \end{split}
\end{equation}
}

\section{Method}
\vspace{-1mm}
We incorporate neural BRDFs into PBR to collaboratively learn complex appearance properties (\textit{i.e.} diffuse and specular albedo) and sophisticated geometry attributes. The proposed \textit{NeuFace} performs 3D face capturing from multi-view images, which only requires additional camera poses. As Fig.~\ref{fig:pipeline} illustrates, it exploits a \textit{Spatial MLP} and an \textit{Integrated Basis MLP} to learn the skin BRDFs and simulates the physically-correct light transport by explicitly decomposing it into diffuse albedo, light integral, and BRDFs integral. Facial appearance modeling can be formulated by Eq.~\eqref{rendering_eq2}, where we make great efforts to tackle the challenges caused by high computational cost and data demand during the learning procedure. For more accurate estimation, \textit{NeuFace} further makes use of an SDF-based representation, \ie \textit{ImFace}~\cite{zheng2022imface}, as the facial geometry prior in forward mapping. The facial appearance and geometry field are jointly recovered by an end-to-end inverse rendering.

\subsection{Specular Modeling}
\label{secspec}

The specular term, \ie $L_{\mathrm{s}}$ with BRDF $f_{\mathrm{s}}(x, \omega_i, \omega_o)$, is intractable to solve due to the view-dependent characteristic of $\omega_o$. In prior studies dealing with non-face neural rendering, the analytic BRDFs, \eg, microfacet models~\cite{munkberg2022extracting, zhang2021physg, srinivasan2021nerv}, have been explored for an approximated integral. Nevertheless, these explicit models can only recover a small set of materials while excluding facial skin which exhibits quite unique properties~\cite{murakami2002measurement}. For a stronger representation ability, NeRFactor~\cite{zhang2021nerfactor} employs the data-driven BRDFs learned from the real-world captured MERL database~\cite{merl}, but the live tissues like facial skin with spatially-varying BRDFs are extremely difficult to measure in vivo~\cite{igarashi2005appearance}. Moreover, heavy Monte Carlo sampling has to be performed in NeRFactor to solve the rendering equation, which is not practical in our case as the facial geometry is originally unknown.
 
In this study, we propose an alternative to render complicated facial highlights without external data and numerical integration. In particular, inspired by the split-integral approximation in real-time rendering~\cite{akenine2019real}, the specular is decomposed, where $L_{\mathrm{s}}$ can be approximated by:
\vspace{-1mm}
{\small
\begin{equation}
    L_{\mathrm{s}}\!\approx\!\varrho\int_{\mathrm{s}^2}  f_{\mathrm{s}}\left(x, \omega_i, \omega_o\right) (\omega_i \cdot \mathbf{n})^{+} \mathrm{d} \omega_i 
    \int_{\mathrm{s}^2} D(h) L_i\left(\omega_i\right) \mathrm{d} \omega_i ,
    \label{splitsum}
\end{equation}
}
\hspace{-0.5em} where $D(h)$ is a distribution function suggesting how light bounces at the half vector $h\!=\!\frac{\omega_i+\omega_o}{\|\omega_i+\omega_o\|_2}$. The entire equation is flanked by the integral terms of material and light, respectively, which are solved individually in the subsequent.

\textbf{Material Integral.} The first split integral term in Eq.~\eqref{splitsum} is only related to the material property, which is parameterized by a learnable network in our method for a higher representation capacity to facial skin. Considering that the 2D observations can only constrain the integrated values, rather than modeling $f_{\mathrm{s}}\left(x, \omega_i, \omega_o\right)$ as in \cite{zhang2021nerfactor,zheng2021compact}, we directly formulate the entire integral as a smoother function $F$:
{\small
\begin{equation}
    \int_{\mathrm{s}^2} f_{\mathrm{s}}\left(x, \omega_i, \omega_o\right) (\omega_i \cdot \mathbf{n})^{+} \mathrm{d} \omega_i
    = F(x,\omega_o,\mathbf{n}).
    \label{eq:para}
\end{equation}
}
\hspace{0.5em} It should be noted that an enormous number of ground-truth samples are needed to achieve robust fitting with MLP for such a spatially-varying 9-variable function, which is not available in practice. Based on the previous studies on face appearance measurements~\cite{ghosh2008practical, weyrich2006analysis}, we put forward a key assumption that all the facial surface positions of an individual share a similar specular structure, and the spatially-varying properties can thus be represented by diverse linear combinations of only a few (low-rank) learnable BRDFs:

\vspace{-2mm}
{\small
\begin{equation}
     f_{\mathrm{s}}\left(x, \omega_i, \omega_o\right) \approx \sum_{j=1}^k {c_j(x) b_j(\omega_i, \omega_o)},
     \label{lowrank1}
\end{equation}
}
\hspace{-0.5em} where $\{b_j(\omega_i, \omega_o)\}_{j=1}^k$ denotes $k$ global space-independent BRDF bases, and $\mathbf{c}(x)=[c_1(x),c_2(x),...,c_k(x)]^T$ means the corresponding linear coefficients for each surface position $x$. As such, the material integral can be formulated as:

\vspace{-2mm}
{\small
\begin{equation}
    \int_{\mathrm{s}^2} f_{\mathrm{s}}\left(x, \omega_i, \omega_o\right) (\omega_i \cdot \mathbf{n})^{+} \mathrm{d} \omega_i
    = \mathbf{c}(x) \cdot \mathbf{B}(\omega_o,\mathbf{n}),
\end{equation}
}
\hspace{-0.5em} where $\mathbf{B}(\omega_o,\mathbf{n})=[B_1,B_2,...,B_k]^T$ indicates $k$ integrated BRDF basis $b_j$ multiplied by half-cosine function $(\omega_i \cdot \mathbf{n})^{+}$:
\vspace{-2mm}
{\small
\begin{equation}
    B_j(\omega_o,\mathbf{n}) = \int_{\mathrm{s}^2} b_j\left(\omega_i, \omega_o\right) (\omega_i \cdot \mathbf{n})^{+} \mathrm{d} \omega_i,  j=1,2,...,k.
\end{equation}
\vspace{-2mm}
}

We leverage an MLP, namely \textit{Integrated Basis MLP}, to fit $\mathbf{B}(\omega_o,\mathbf{n})$. $\omega_o\cdot\mathbf{n}$ is also fed into it to account for the Fresnel effects. Meanwhile, coefficient vector $\mathbf{c}(x)$ is predicted by a \textit{Spatial MLP} with $x$ as input. Note that facial skin is dielectric and cannot color the highlight, we thus use a single channel $B_j$ to represent the monochromatic facial BRDFs.

As depicted in Fig.~\ref{fig:pipeline}~(c), the low-rank prior globally restricts the solution space for all facial specularities without enforcing any spatial smoothness~\cite{yao2022neilf} or clustering of sampling positions \cite{munkberg2022extracting}. With such a prior, the material integral term is much easier to fit and interpolate, producing impressive results with moderate training data.

\textbf{Light Integral.} For glossy surfaces, the reflected energy is mostly concentrated near the reflection direction, denoted by $\omega_r$, as illustrated in Fig.~\ref{fig:pipeline}~(b). Accordingly, we model $D(h)$ with a 3-dimensional normalized Spherical Gaussian distribution ($\operatorname{vMF}$), centered at $\omega_r$ with a concentration parameter $\kappa$ indicating the shininess:
{\small
\begin{equation}
    D(h) \approx \operatorname{vMF}(\omega_i;\omega_r,\kappa),    
\end{equation}
}
\hspace{-0.5em} $\kappa$ is predicted by \textit{Spatial MLP}, and a larger value suggests a sharper underlying BRDF lobe pointing to $\omega_r$, making facial skin look shinier. As the proof in our \textit{Supplementary Material} shows, the second split integral in Eq.~\eqref{splitsum} which describes specular light transport can be approximated by:
{\small
\begin{equation}
     \int_{\mathrm{s}^2} D(h) L_i\left(\omega_i\right) \mathrm{d} \omega_i \approx \sum_{\ell=0} \sum_{m=-\ell}^\ell e^{-\frac{\ell(\ell+1)}{2\kappa}} c_{\ell m} {\mit Y}_{\ell m}(\omega_r).
     \label{spec_light}
\end{equation}
}
\hspace{0.5em} The entire specular term in the neural BRDFs (Eq.~\eqref{rendering_eq2}) can then be efficiently and differentiably computed as:
{\small
\begin{equation}
    L_{\mathrm{s}}\!\approx\!\varrho \left(\mathbf{c}(x) \cdot \mathbf{B}(\omega_o , \mathbf{n})\right) \sum_{\ell=0} \sum_{m=-\ell}^\ell e^{-\frac{\ell(\ell+1)}{2\kappa}} c_{\ell m} {\mit Y}_{\ell m}(\omega_r).
    \label{spec_final}
\end{equation}
}

\subsection{Diffuse Modeling}
\label{secdiff}

Based on Eq.~\eqref{sh_l}, diffuse radiance $L_\mathrm{d}$ at surface position $x$ can be rewritten as:

\vspace{-2mm}
{\small
\begin{equation}
    L_{\mathrm{d}}(x)=\frac{{\bf a}(x)}{\pi} \sum_{\ell=0} \sum_{m=-\ell}^\ell c_{\ell m} \int_{\mathrm{s}^2} Y_{\ell m}\left(\omega_i\right)\left(\omega_i \cdot \mathbf{n}\right)^{+} \mathrm{d} \omega_i .
\end{equation}
}
\hspace{0.5em} Following the Funk-Hecke theorem~\cite{basri2003lambertian}, the convolution of $(\omega_i \cdot \mathbf{n})^{+}$ with spherical harmonics $Y_{\ell m}$ can be analytically calculated by:

\vspace{-1mm}
{\small
\begin{equation}
    \int_{\mathrm{s}^2} Y_{\ell m}\left(\omega_i\right)\left(\omega_i \cdot \mathbf{n}\right)^{+} \mathrm{d} \omega_i
    \approx
    {\boldsymbol{\Lambda}_{\ell m}} Y_{\ell m}(\mathbf{n}) ,
\end{equation}
}
\hspace{-0.5em} where

\vspace{-1mm}
{\small
\begin{equation}
    {\boldsymbol{\Lambda}_{\ell m}}=
    \begin{cases} 
    \frac{2 \pi}{3}, & \text{if }\ell=1 , \\
    \frac{(-1)^{{\ell}/{2}+1} \pi}{2^{\ell-1}(\ell-1)(\ell+2)} \tbinom{\ell}{\ell/2}, &  \text{if } \ell \text { is even } , \\
    0, &  \text{if } \ell \text { is odd } ,
    \end{cases}
\end{equation}
}
\hspace{-0.5em} Refer to \cite{basri2003lambertian} for details. In \textit{NeuFace}, diffuse albedo ${\bf a}(x)$ is modeled by \textit{Spatial MLP}. With learnable coefficients $c_{lm}$ of SH lighting, the diffuse term can be directly computed as:

\vspace{-1mm}
{\small
\begin{equation}
    L_{\mathrm{d}}(x) \approx \frac{{\bf a}(x)}{\pi} \sum_{\ell=0} \sum_{m=-\ell}^\ell {\boldsymbol{\Lambda}_{\ell m}} c_{\ell m} Y_{\ell m}\left(\mathbf{n}\right) .
    \label{diff_final}
\end{equation}
}
\hspace{0.5em} In summary, the appearance component of \textit{NeuFace} is composed of a \textit{Spatial MLP}$: x \mapsto (\varrho, \mathbf{c}, \kappa, {\bf a})$, an \textit{Integrated Basis MLP}$: (\omega_o,  \mathbf{n}, \omega_o\cdot\mathbf{n}) \mapsto \mathbf{B}$, and the learnable environment light coefficients, \ie $c_{\ell m}$. Radiance $L_o$ can finally be estimated through Eq.~\eqref{spec_final} and Eq.~\eqref{diff_final}.

\subsection{Geometry Modeling}
\label{secgeo}

To achieve end-to-end training, a differentiable geometry representation is essential. Similar to most neural rendering practices \cite{zhang2021physg, wang2021neus, yariv2020idr, yariv2021volsdf}, a neural SDF can be used to implicitly define the facial geometry. Here, we leverage \textit{ImFace}~\cite{zheng2022imface} as a direct facial SDF prior to facilitate sampling and training. To capture the geometries outside the prior distribution, we fine-tune \textit{ImFace} $\mathcal{I}:x \mapsto \text{SDF}$ and introduce a neural displacement field $\mathcal{D}(x)$ to correct the final results:
{\small
\begin{equation}
    \text{SDF}(x)=\mathcal{I}(x)+\mathcal{D}(x).
\end{equation}
}
\hspace{0.5em} Based on the property of SDF, surface normal $\mathbf{n}$ can be extracted by auto-gradient: $\mathbf{n}=\mathbf{\nabla} \text{SDF}(x)$.

\begin{figure}
  \centering
  \setlength{\abovecaptionskip}{0pt}
  \setlength{\belowcaptionskip}{4pt}
  \includegraphics[width=0.9\linewidth]{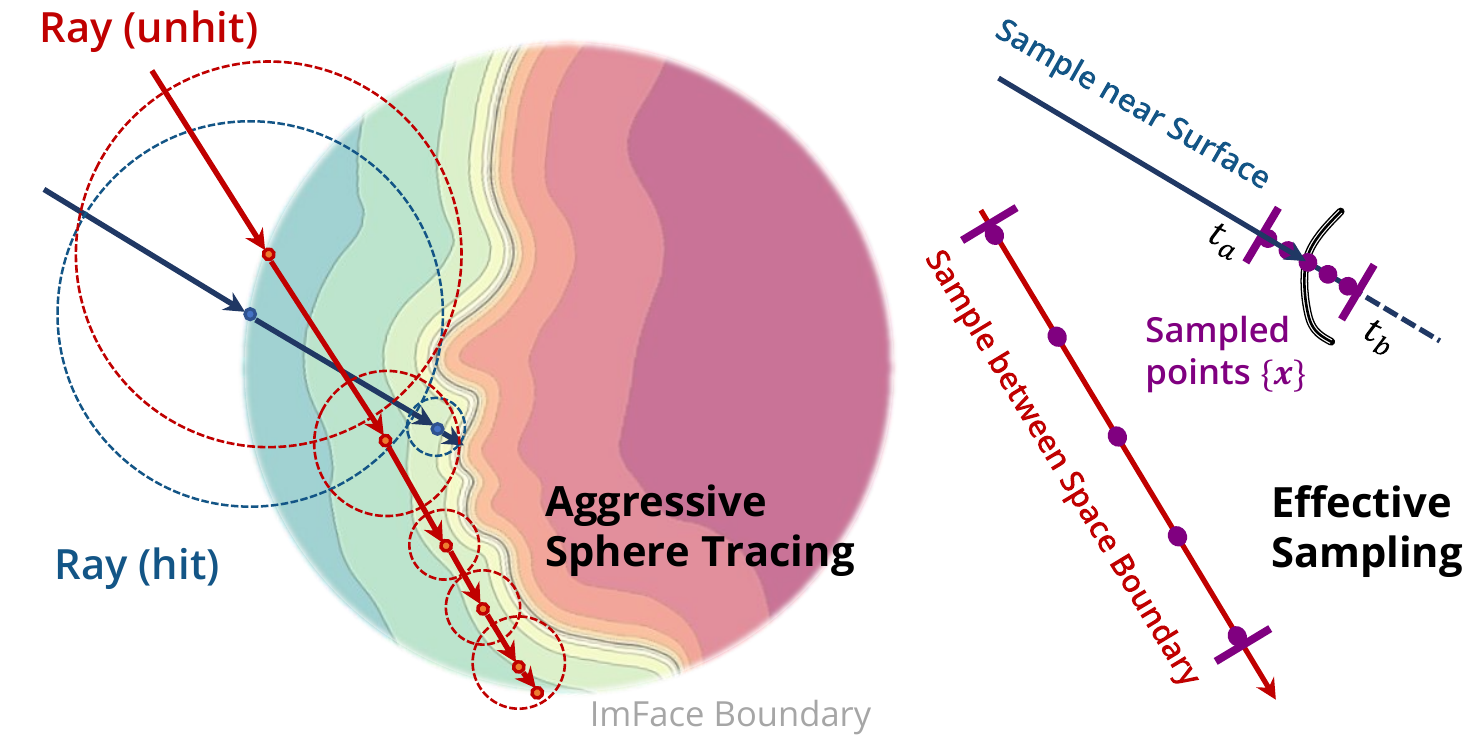}
   \caption{Illustration of the sampling strategy. Aggressive sphere tracing is performed to accelerate the sampling process for volume rendering. For the rays unhit the surface, the sphere defined by the geometry prior of \textit{ImFace} is used to constrain the sampling space.}
   \label{fig:VolumeRendering}
   \vspace{-8mm}
\end{figure}

\subsection{Sampling and Rendering}
\label{secforward}

In order to cope with multi-layered facial skin, we volumetrically render the radiance value as in VolSDF~\cite{yariv2021volsdf}. For a ray $x(t)$ emanated from a camera position $\mathbf{o} \in \mathbb{R}^3$ in direction $\omega_o$, defined by $x(t)=\mathbf{o}+t \omega_o, t>0$, the density function is defined as $\sigma(t)=\beta^{-1} \Psi_\beta (-\text{SDF}(x(t)))$. $\Psi_\beta$ is the cumulative distribution function of the Laplace distribution with zero mean and learned scale parameters $\beta$. As such, the integrated radiance for each ray is evaluated by:
\vspace{-2mm}
{\small
\begin{equation}
    \mathbf{I}(\mathbf{o},\omega_o)=\int_{t_a}^{t_b} {L_o(x(t),\omega_o) \sigma(t) T(t) }\mathrm{d} t,
\end{equation}
}
\hspace{-0.5em} where $T(t)=\text{exp}({-\int_{0}^t \sigma(s) \mathrm{d} s })$ is the transparency.

For rendering acceleration, instead of densely sampling points along the ray as in \cite{yariv2021volsdf}, we first perform aggressive sphere tracing~\cite{liu2020dist} to quickly find position $t_0$ near the surface (with a 0.05$mm$ threshold), and 32 points are then uniformly sampled from $t \in [t_a,t_b]$, where $t_a=t_0-0.5mm$, $t_b=t_0+0.5mm$. As Fig.~\ref{fig:VolumeRendering} shows, for the rays that do not hit the surface, we sparsely sample points within sphere $\Omega$ defined by the geometry prior from \textit{ImFace}~\cite{zheng2022imface} and calculate the accumulation, bypassing the mask loss~\cite{yariv2020idr} which requires heavy sampling. By combining volume and surface rendering, it makes a good trade-off balancing the rendering quality, geometry precision, and sampling efficiency. 

\textbf{Neural Photometric Calibration.} To auto-calibrate the inconsistent color response and white balance among cameras, we apply per-image linear mapping matrices $\mathbf{A}_n \in \mathbb{R}^{3\times3}$ to the rendered radiance values:

\vspace{-2mm}
{\small
\begin{equation}
    \mathbf{I}_{n}=\mathbf{A}_n \mathbf{I},
\end{equation}
}
\hspace{-0.39em}where $\mathbf{A}_n$ is predicted by a lightweight MLP conditioned on learnable per-image embedding.

\begin{figure*}
  \centering
  \includegraphics[width=1\linewidth]{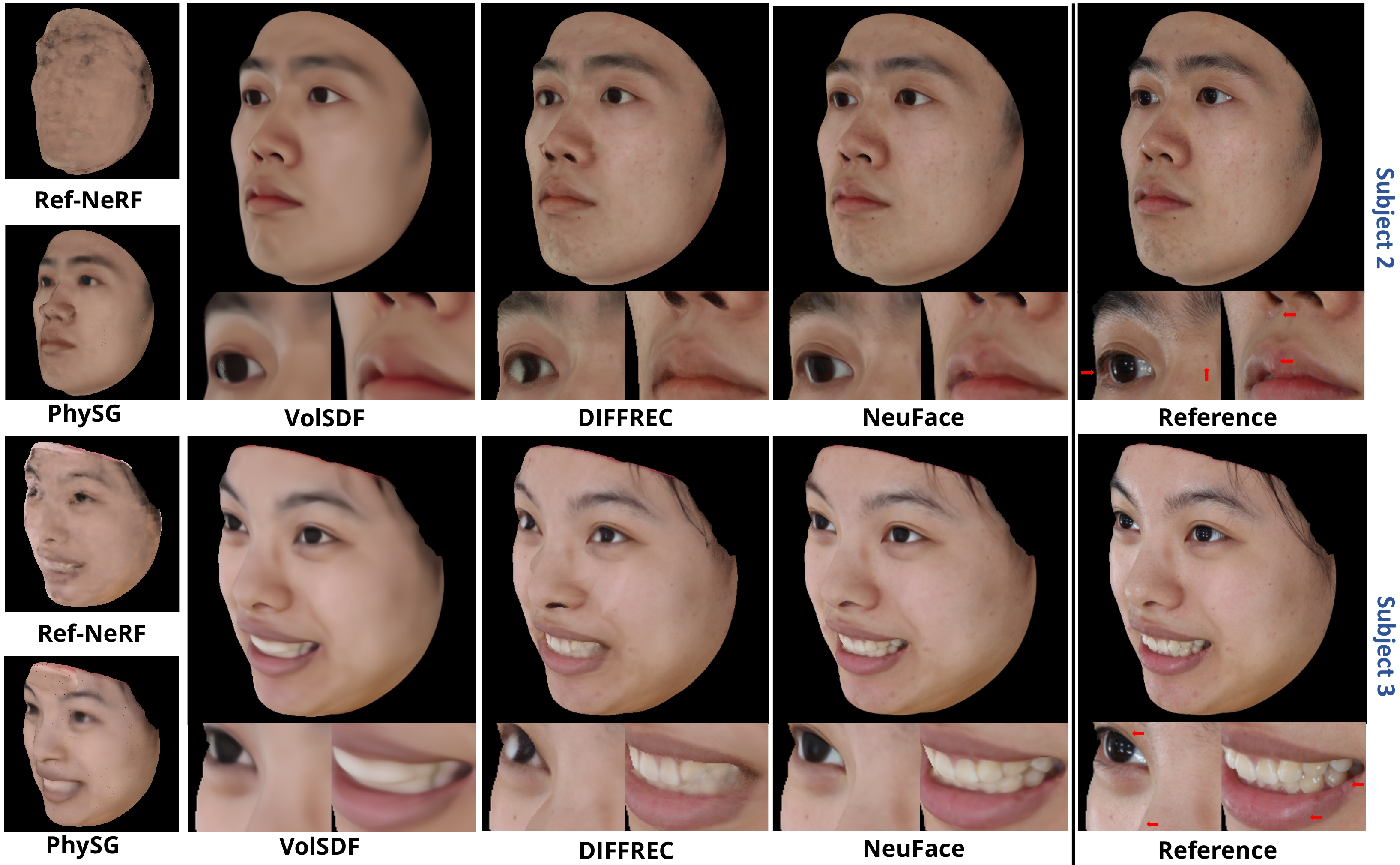}
  \vspace{-6mm}
  \caption{Comparison of novel view synthesis with Ref-NeRF\cite{verbin2022ref}, PhySG\cite{zhang2021physg}, VolSDF\cite{yariv2021volsdf} and DIFFREC\cite{munkberg2022extracting}. Zoom in for a better view. \textit{NeuFace} captures faces with much richer skin details and high-fidelity highlights. }
  \label{fig:main_result}
\end{figure*}

\subsection{Loss Functions}
\label{secloss}

\textit{NeuFace} is trained with a compound critic to learn accurate appearance and geometry properties:

\vspace{-2mm}
{\small
\begin{equation}
  {\cal L}\!=\! {\cal L}_{RGB}+{\cal L}_{white}+ {\cal L}_{spec}+{\cal L}_{geo}.
\end{equation}}
\hspace{0.5em} Let $P_n$ be the pixels of the $n$-th multi-view image, ${\bar {\mathbf{I}}}_n$. Each term of ${\cal L}$ is described as follows, where $\lambda$ denotes the trade-off hyper-parameter.

\para{Reconstruction Loss} It is used to supervise the rendered 2D faces being close to the real observations:
{\small
\begin{equation}
  {\cal L}_{RGB}\!=\!{\lambda_1} \sum_{n} \sum_{\omega_o \in P_n} |\mathbf{I}_{n}-{\bar {\mathbf{I}}}_n|.
\end{equation}
}

\para{Light Regularization}
We assume a nearly white environment light in capturing, achieved by:
\vspace{-2mm}
{\small
\begin{equation}
  {\cal L}_{white}\!=\!{\lambda_2} \sum_{n} \sum_{\omega_o \in P_n} |L_i-{\bar L}_i|,
  \vspace{-2mm}
\end{equation}
}
\hspace{-0.5em} where ${\bar L}_i$ is calculated by averaging RGB channels.

\para{Facial Specular Regularization} Only a small ratio of the incident light on facial surfaces (roughly 6\%~\cite{valery2000tissue}) reflects directly, and we thus penalize the specular energy by:
\vspace{-2mm}
{\small
\begin{equation}
  {\cal L}_{spec}\!=\!{\lambda_3} \sum_{n}  \sum_{\omega_o  \in P_n}
  L_s.
  \vspace{-2mm}
\end{equation}
}

\para{Geometry Losses} Following \cite{zheng2022imface}, the embedding regularization, Eikonal regularization, and a new residual constraint are exploited for accurate geometry modeling:
\vspace{-2mm}
{\small
\begin{equation}
  {\cal L}_{geo}\!=\!{\lambda_4} {\|{\mathbf z}\|}^2 \!+\!
   {\lambda_5 } \sum_{x \in \Omega} |\|\nabla \text{SDF}(x)\|-1| \!+\!
  {\lambda_6} \sum_{x \in \Omega} |\mathcal{D}(x)|
  ,
  \vspace{-2mm}
\end{equation}
}
\hspace{-0.5em} where ${\mathbf z}$ represents the embeddings of \textit{ImFace}~\cite{zheng2022imface}.

\section{Experiments}

We extensively conduct experiments for both subjective and objective evaluation and compare our \textit{NeuFace} with the state-of-the-art neural rendering methods. Ablation studies are performed to validate the specifically designed modules. For more implementation details on the network architecture and training procedure, please refer to the \textit{Supplement}.

FaceScape \cite{yang2020facescape} is adopted, and it is a large-scale 3D face dataset including high-quality multi-view images of 4K resolution, which is collected from 359 real subjects with 20 expressions. During capturing, around 58 cameras with different types and settings are equally distributed on a spherical structure with uniform illumination. We use the authorized data from 3 individuals for model evaluation, with 43 images for training and 11 images for testing. The images are downsampled to 1K resolution.

\begin{table*}[t]
    \begin{threeparttable}
    \resizebox{1.\textwidth}{!}{
    \begin{tabular}{l|c|c|cccc|cccc|cccc}
    {\multirow{2}{*}{\makebox[0.007\textwidth][l]{\small w/ our calibration}}} 
    & {\multirow{2}{*}{\makebox[0.002\textwidth][c]{\scriptsize F.}}} 
    & {\multirow{2}{*}{\makebox[0.003\textwidth][c]{\scriptsize S.N. $\downarrow$}}} 
    & \multicolumn{4}{c|}{\textbf{Subject 1 (Rich Reflection)}}
    & \multicolumn{4}{c|}{\textbf{Subject 2 (Moderate Reflection)}}
    & \multicolumn{4}{c}{\textbf{Subject 3 (Low Reflection)}} \\
    & & & \multicolumn{1}{c}{\scriptsize PSNR $\uparrow$} 
        & \multicolumn{1}{c}{\scriptsize SSIM $\uparrow$}
        & \multicolumn{1}{c}{\scriptsize LPIPS $\downarrow$}
        & \multicolumn{1}{c|}{\scriptsize Chamfer $\downarrow$} 
    & \multicolumn{1}{c}{\scriptsize PSNR $\uparrow$} 
    & \multicolumn{1}{c}{\scriptsize SSIM $\uparrow$}
    & \multicolumn{1}{c}{\scriptsize LPIPS $\downarrow$}
    & \multicolumn{1}{c|}{\scriptsize Chamfer $\downarrow$} 
    & \multicolumn{1}{c}{\scriptsize PSNR $\uparrow$} 
    & \multicolumn{1}{c}{\scriptsize SSIM $\uparrow$}
    & \multicolumn{1}{c}{\scriptsize LPIPS $\downarrow$}
    & \multicolumn{1}{c}{\scriptsize Chamfer $\downarrow$} \\
    \shline
     NeRF\cite{mildenhall2020nerf} & \ding{55} & 256 & \gc{13.98} & \gc{0.712} & \gc{.1378} & \gc{3.666} & \gc{13.67} & \gc{0.724} & \gc{.1493} & \gc{1.910} & \gc{-} & \gc{-} & \gc{-} & \gc{-} \\
     Ref-NeRF\cite{verbin2022ref} & \ding{55} & 256 & {19.83} & {0.845} & {.1134} & {2.042} & {20.01} & {0.846} & {.1410} & {1.782} & {17.33} & {0.820} & {.1381} & {3.682} \\
     VolSDF\cite{yariv2021volsdf} & \ding{55} & 610 & \cellcolor{orange}{29.03} & \cellcolor{lightyellow}{0.941} & \cellcolor{lightyellow}{.0502} & \cellcolor{orange}{0.516} & \cellcolor{orange}{28.40} & \cellcolor{orange}{0.937} & \cellcolor{lightyellow}{.0730} & \cellcolor{orange}{0.589} & \cellcolor{lightyellow}{26.74} & \cellcolor{lightyellow}{0.926} & \cellcolor{lightyellow}{.0580} & \cellcolor{orange}{0.626} \\
     PhySG\cite{zhang2021physg} & \ding{52} & 40.2 & \cellcolor{lightyellow}{27.77} & {0.928} & {.0611} & {0.864} & {26.36} & {0.913} & {.0864} & {0.977} & {23.07} & {0.902}  & {.0735} & {1.776}\\
     DIFFREC*\cite{munkberg2022extracting} & \ding{52} & 39.5 & {27.72} & \cellcolor{orange}{0.942} & \cellcolor{orange}{.0244} & \cellcolor{lightyellow}{0.546} & \cellcolor{lightyellow}{27.61} & \cellcolor{lightyellow}{0.934} & \cellcolor{orange}{.0353} & \cellcolor{lightyellow}{0.606} & \cellcolor{orange}{27.02} & \cellcolor{orange}{0.930} & \cellcolor{orange}{.0372} & \cellcolor{lightyellow}{0.727} \\
     \hline
     \textit{NeuFace} & \ding{52} & {35.8} & \cellcolor{tablered}\textbf{31.25} &  \cellcolor{tablered}\textbf{0.958} &  \cellcolor{tablered}\textbf{.0237} &  \cellcolor{tablered}\textbf{0.447} &  \cellcolor{tablered}\textbf{31.87} &  \cellcolor{tablered}\textbf{0.961} &  \cellcolor{tablered}\textbf{.0314} &  \cellcolor{tablered}\textbf{0.481} &  \cellcolor{tablered}\textbf{30.92} &  \cellcolor{tablered}\textbf{0.953} &  \cellcolor{tablered}\textbf{.0240} &  \cellcolor{tablered}\textbf{0.533} \\
    \end{tabular}}

    \begin{tablenotes}
    \item \hspace{-1.8em} {\scriptsize `F.' refers to factorization. `S.N.' refers to the number of sampled points per ray. DIFFREC* is implemented with neural surface. NeRF fails to converge on Subject 3.}
    \end{tablenotes}
    
    \end{threeparttable}
    \caption{Quantitative comparison with NeRF\cite{mildenhall2020nerf}, Ref-NeRF\cite{verbin2022ref}, VolSDF\cite{yariv2021volsdf}, PhySG\cite{zhang2021physg}, and DIFFREC\cite{munkberg2022extracting}. \textit{NeuFace} outperforms the counterparts by a large margin in both appearance and geometry metrics. Red, orange, and yellow: the best, second-best, and third-best.
    \label{tab:main_result}
    }
    \vspace{-4mm}
\end{table*}

\subsection{Comparison} 
 
To validate the ability of \textit{NeuFace} for simultaneous appearance and geometry capturing, we compare it with the state-of-the-art approaches to view synthesis~\cite{mildenhall2020nerf, verbin2022ref}, geometry reconstruction~\cite{yariv2021volsdf} and appearance factorization~\cite{zhang2021physg, munkberg2022extracting}. Considering that neural rendering is extremely sensitive to camera inconsistency, we apply the proposed neural photometric calibration to all the counterparts for fair comparison. Because DIFFREC~\cite{munkberg2022extracting} fails to reconstruct all the faces with explicit tetrahedral grids, we implement a more stable neural surface version instead. PSNR, SSIM~\cite{wang2004image} and LPIPS~\cite{zhang2018unreasonable} are used as the metrics to evaluate the view synthesis quality, and the Chamfer distance to the ground-truth 3D model is used to measure the geometry accuracy.  

\begin{figure}
  \centering
  \setlength{\abovecaptionskip}{0pt}
  \setlength{\belowcaptionskip}{6pt}
  \includegraphics[width=1.\linewidth]{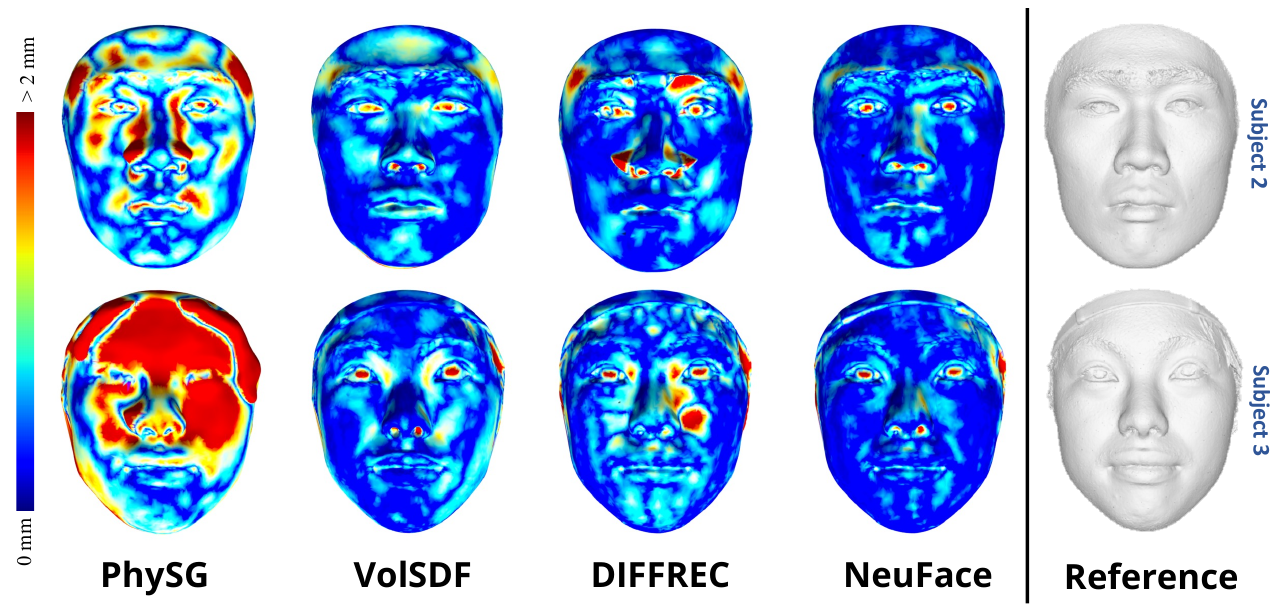}
   \caption{Comparison of geometry reconstruction to PhySG\cite{zhang2021physg}, VolSDF\cite{yariv2021volsdf}, and DIFFREC\cite{munkberg2022extracting}. \textit{NeuFace} achieves lower errors.}
   \label{fig:geometry}
\end{figure}

\textbf{Novel View Synthesis.} Fig.~\ref{fig:main_result} displays the results. It can be seen that as a recent advanced view-synthesis method, Ref-NeRF~\cite{verbin2022ref} fails to recover facial appearances, mainly due to the strong shape-appearance ambiguity on faces. Designed for geometry reconstruction, VolSDF~\cite{yariv2021volsdf} sacrifices the appearance quality, resulting in blurred face renderings. PhySG~\cite{zhang2021physg} and DIFFREC~\cite{munkberg2022extracting} are also physically-based neural rendering methods as ours, which aim at end-to-end appearance and geometry recovery. However, limited by the inadequate capacity of the analytic BRDFs, they can only recover rough facial appearances with obvious artifacts. By contrast, \textit{NeuFace} achieves realistic face appearances with much richer skin details and high-fidelity highlights, \eg, the areas around lips, glabellar, and nose. The quantitative results in Tab.~\ref{tab:main_result} confirm its effectiveness,
which are evaluated on smile subjects 1, neutral subjects 2 and 3.

\begin{table}
    \centering
    \resizebox{.48\textwidth}{!}{
    \begin{tabular}{@{}l|cccc}
    Shading Model & PSNR $\uparrow$ & SSIM $\uparrow$ & LPIPS $\downarrow$ & Chamfer $\downarrow$ \\
    \shline
    SH+Phong Model & 29.31 & 0.951 & .0300 & 0.556 \\
    Cubemap+Disney BRDF & \cellcolor{lightyellow}{30.45} & \cellcolor{lightyellow}{0.954} & \cellcolor{orange}{.0242} & \cellcolor{lightyellow}0.524  \\
    \hline
    Cubemap+Neural Bases & \cellcolor{orange}30.82 & \cellcolor{tablered}\textbf{0.958} & \cellcolor{lightyellow}.0256 &  \cellcolor{tablered}\textbf{0.426} \\
    SH+Neural Bases (Ours) &  \cellcolor{tablered}\textbf{31.25} &  \cellcolor{tablered}\textbf{0.958} &   \cellcolor{tablered}\textbf{.0237} &  \cellcolor{orange}{0.447} \\
    \end{tabular}} 
    \caption{Quantitative ablation study on the shading model.}
    \label{table:quantitative}
\end{table}

\textbf{Geometry Reconstruction.} Fig.~\ref{fig:geometry} makes comparison of geometry reconstruction with PhySG~\cite{zhang2021physg}, VolSDF~\cite{yariv2021volsdf}, DIFFREC~\cite{munkberg2022extracting}, and ground-truth, measured by color-coded distance. \textit{NeuFace} reports the lowest errors and even outperforms the geometry-oriented method, \ie VolSDF, suggesting that our PBR framework benefits not only facial appearances but also facial shapes. The quantitative results in Tab.~\ref{tab:main_result} further validate this claim. Tab.~\ref{tab:main_result} also demonstrates the efficiency of our rendering strategy (Section~\ref{secforward}). Our method requires fewer sampling points than general volume rendering and is on par with surface rendering, which needs additional sampling to minimize the mask loss.

\begin{figure}
  \centering
  \setlength{\abovecaptionskip}{0pt}
  \setlength{\belowcaptionskip}{0pt}
  \includegraphics[width=1.\linewidth]{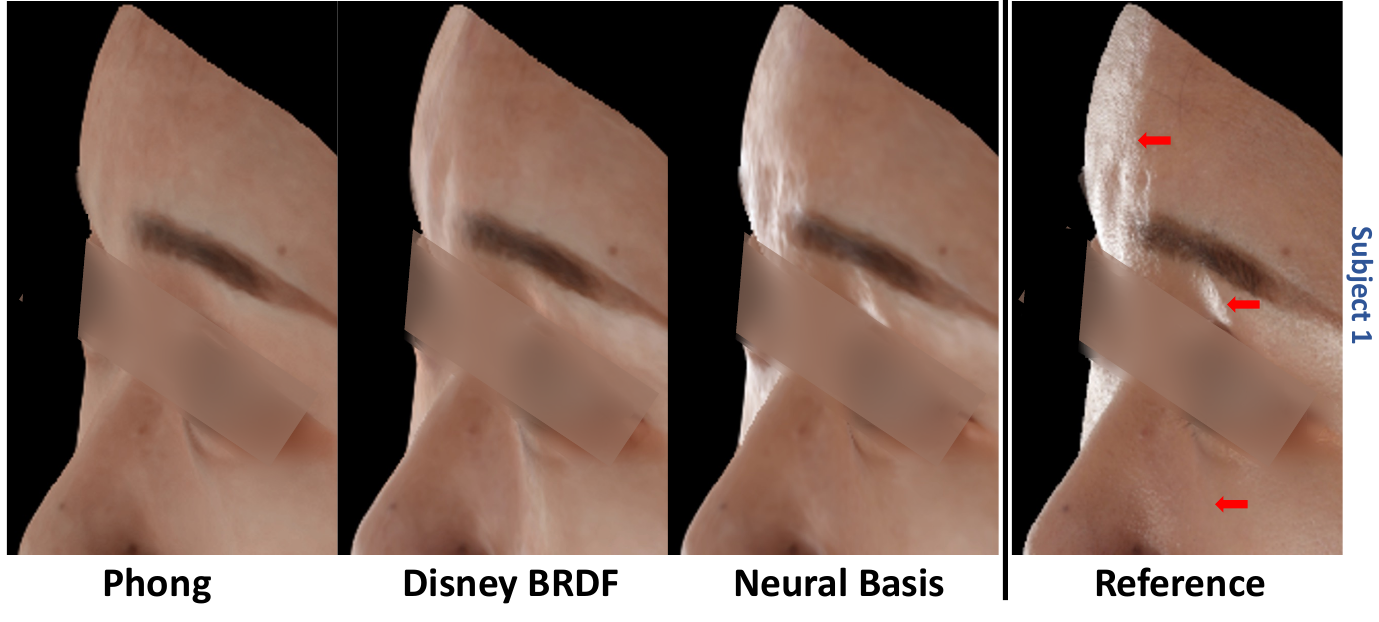}
   \caption{Comparison with the Phong~\cite{phong1975illumination} and Disney BRDFs~\cite{burley2012physically}.}
   \label{fig:ablation_brdf}
\end{figure}

\begin{figure*}
  \centering
  \setlength{\abovecaptionskip}{0pt}
  \setlength{\belowcaptionskip}{0pt}
  \includegraphics[width=1.0\linewidth]{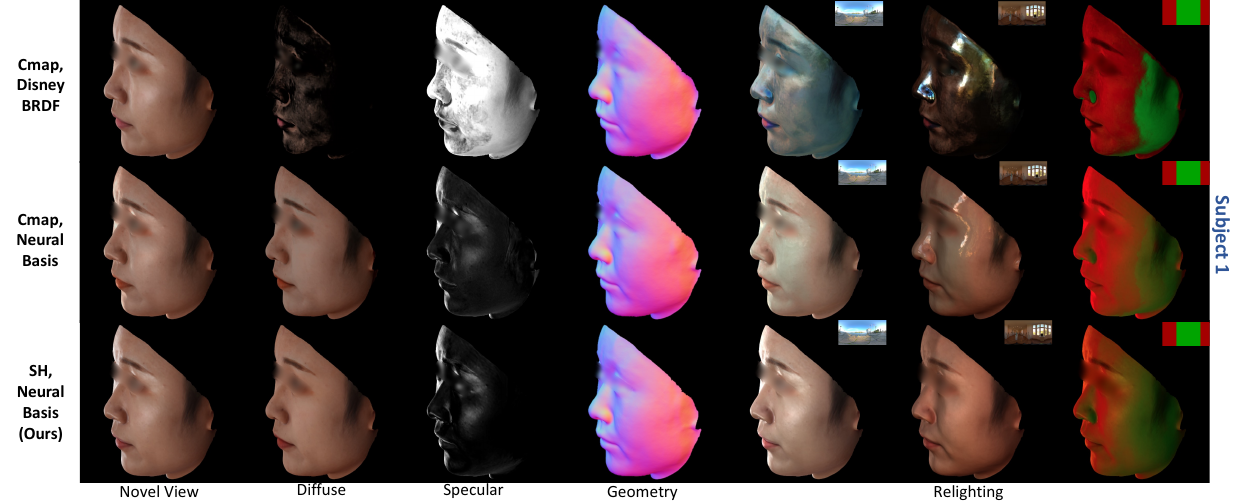}
    \caption{Ablation on shading models. Ours (bottom) compares with improved DIFFREC~\cite{munkberg2022extracting} (upper) and Cubemap lighting (middle).}
   \label{fig:factorization2}
\end{figure*}

\begin{table}
    \centering
    \resizebox{.48\textwidth}{!}{
    \begin{tabular}{@{}l|cccc}
    Designs & PSNR $\uparrow$ & SSIM $\uparrow$ & LPIPS $\downarrow$ & Chamfer $\downarrow$ \\
    \shline
    w/o low rank (directly fitting) & \cellcolor{orange}{29.45} & \cellcolor{lightyellow}{0.950} & \cellcolor{lightyellow}{.0304} & \cellcolor{lightyellow}{0.553}  \\
    w/o low rank (256-D feature) & \cellcolor{lightyellow}{29.43} & \cellcolor{orange}{0.951} & \cellcolor{orange}{.0302} &  \cellcolor{orange}{0.544} \\
    \hline
    \textit{NeuFace} &  \cellcolor{tablered}\textbf{31.25} &  \cellcolor{tablered}\textbf{0.958} &   \cellcolor{tablered}\textbf{.0237} &  \cellcolor{tablered}\textbf{0.447} \\
    \end{tabular}} 
    \caption{Quantitative ablation study on the low-rank prior.}
    \label{table:quantitative2}
\end{table}

\subsection{Ablation Study}

The following experiments are performed on Subject 1 as the face images exhibit more complex highlights.  

\textbf{On Shading Model.} Our neural BRDFs module is designed for a higher representation capability to facial skin, and we carry out an ablation study by replacing it with the well-reputed Phong~\cite{phong1975illumination} and Disney BRDFs~\cite{burley2012physically}. In particular, the specular term in the Phong BRDFs is approximated as in \cite{ramamoorthi2002frequency}, controlled by a single shininess parameter. The Disney BRDFs are implemented by using the code provided by DIFFREC~\cite{munkberg2022extracting}. We show novel view rendering results in Fig.~\ref{fig:ablation_brdf}, which demonstrate that the proposed neural BRDFs module can capture high-fidelity reflections of facial skin, such as the sheen around grazing angles, while such light transport cannot be well modeled by the Phong and Disney BRDFs. Tab.~\ref{table:quantitative} also reveals the higher representation ability of neural BRDFs in learning sophisticated skin reflections.

Moreover, we show factorization and relighting results in Fig.~\ref{fig:factorization2}, where SH lighting is compared with Cubemap. As shown, Cubemap indeed achieves comparable results of novel view synthesis and even slightly better geometries by combining our other designs. However, in our under-constrained settings, it fails to accurately relight faces under new environment maps. In contrast, \textit{NeuFace} can well adapt to arbitrary lighting conditions.

\textbf{On Neural Integrated Basis.} The low-rank prior is introduced to constrain the solution space and learn accurate neural BRDFs with moderate observations. We compare it with two variants: 1) directly fitting the 9-variable material integral function formulated in Eq.~\eqref{eq:para} (“directly fitting”); and 2) following IDR~\cite{yariv2020idr}, we use \textit{Spatial MLP} to predict a 256-dimensional feature and then feed it into \textit{Integrated Basis MLP} to learn the material integral (“256-D feature”). Tab.~\ref{table:quantitative2} shows that without low-rank regularization, both variants perform significantly worse. Besides, as shown in our \textit{Supplement}, the basis number $k=3$ yields the best scores. Increasing the basis number makes the underlying BRDFs more difficult to fit with limited observations, leading to performance declines for both metrics.

Ablation studies on neural photometric calibration, \textit{ImFace} prior, and the specular energy loss are in the \textit{Supp}.

\subsection{Extension to Common Objects}

\begin{figure}
  \centering
  \setlength{\abovecaptionskip}{0pt}
  \setlength{\belowcaptionskip}{0pt}
  \includegraphics[width=1.\linewidth]{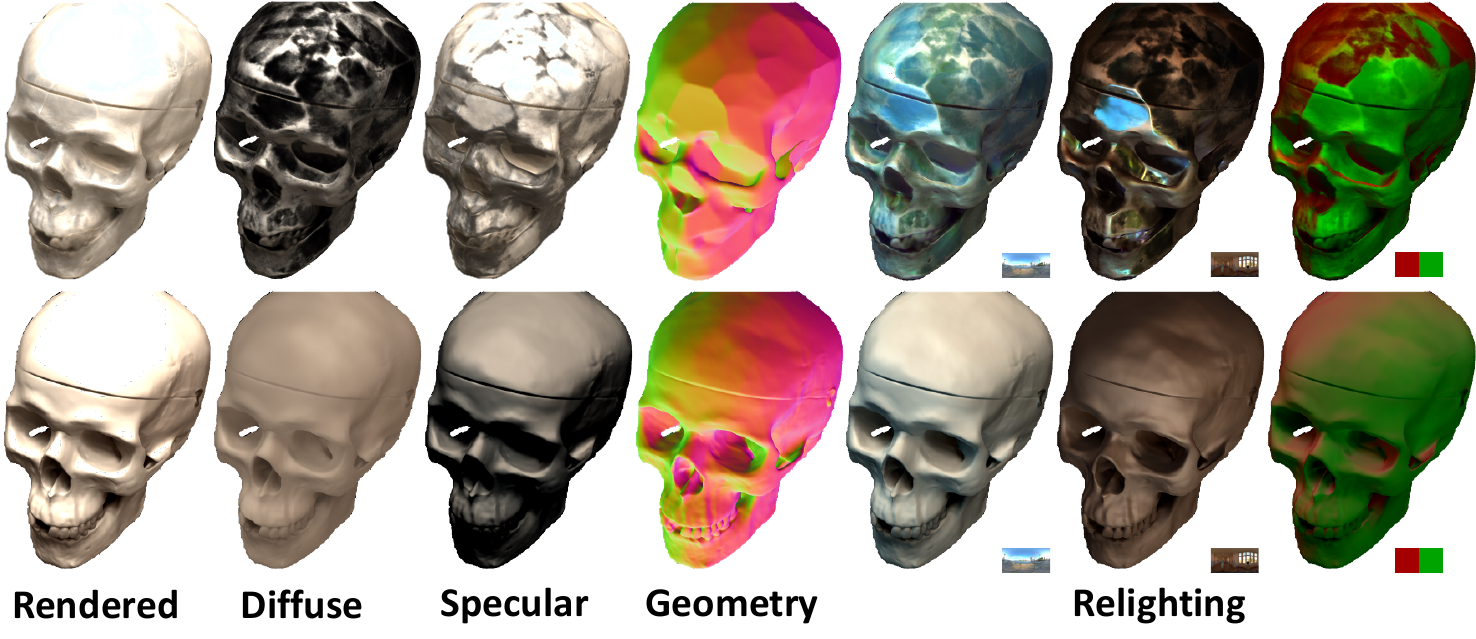}
   \caption{Comparison with DIFFREC on the DTU database~\cite{jensen2014large}. Upper row: DIFFREC; bottom row: ours.}
   \label{fig:dtu}
\end{figure}

To validate the generalization ability of \textit{NeuFace}, we extend it to common objects in the DTU database~\cite{jensen2014large}, which contains large-scale real-world multi-view captures. In this evaluation, the classical IDR~\cite{yariv2020idr} is modified with our rendering pipeline. As Fig.~\ref{fig:dtu} shows, our model achieves more impressive reconstruction and factorization qualities than DIFFREC~\cite{munkberg2022extracting}. More results are in the \textit{Supp}.

\section{Limitation}

The fidelity of the appealing rendering results achieved in this study can be further improved since \textit{NeuFace} mainly focuses on reflection modeling of complex facial skin without explicitly tackling the more challenging issue of subsurface scattering, and only applies a simplified shading model instead. Moreover, \textit{NeuFace} currently delivers a static 3D face rather than a drivable one, and although the geometry model, \emph{i.e.} \textit{ImFace}, is a non-linear morphable model and theoretically supports controllable expression editing, more work is needed for decent performance.

\section{Conclusion}

This paper presents a novel 3D neural rendering model, \textit{i.e.} \textit{NeuFace}, to simultaneously capture realistic facial appearances and geometries from only multi-view images. To cope with complex facial skin reflectance, we bond physically based rendering with neural BRDFs for more accurate and physically-meaningful 3D representations. Moreover, to facilitate the optimization of the underlying BRDFs, we introduce a split-integral technique as well as a simple yet new low-rank prior, which significantly improve the recovering performance. Extensive experiments demonstrate the superiority of \textit{NeuFace} in face rendering and its decent generalization ability to common objects.

\vspace{-3mm}
\section*{Acknowledgments}
\vspace{-1mm}
This work is partly supported by the National Natural Science Foundation of China (No. 62176012, 62202031, 62022011), the Beijing Natural Science Foundation (No. 4222049), the Research Program of State Key Laboratory of Software Development Environment (SKLSDE-2021ZX-04), and the Fundamental Research Funds for the Central Universities.

{\small
\bibliographystyle{ieee_fullname}
\bibliography{egbib}

\begin{thebibliography}{10}\itemsep=-1pt

\bibitem{akenine2019real}
Tomas Akenine-Moller, Eric Haines, and Naty Hoffman.
\newblock {\em Real-time rendering}.
\newblock AK Peters/crc Press, 2019.

\bibitem{basri2003lambertian}
Ronen Basri and David~W Jacobs.
\newblock Lambertian reflectance and linear subspaces.
\newblock {\em IEEE TPAMI}, 25(2):218--233, 2003.

\bibitem{beeler2010high}
Thabo Beeler, Bernd Bickel, Paul Beardsley, Bob Sumner, and Markus Gross.
\newblock High-quality single-shot capture of facial geometry.
\newblock In {\em ACM SIGGRAPH}. 2010.

\bibitem{bradley2010high}
Derek Bradley, Wolfgang Heidrich, Tiberiu Popa, and Alla Sheffer.
\newblock High resolution passive facial performance capture.
\newblock In {\em ACM SIGGRAPH}, 2010.

\bibitem{burley2012physically}
Brent Burley and Walt Disney~Animation Studios.
\newblock {Physically-based shading at Disney}.
\newblock In {\em ACM SIGGRAPH}, 2012.

\bibitem{chen2019learning}
Wenzheng Chen, Huan Ling, Jun Gao, Edward Smith, Jaakko Lehtinen, Alec
  Jacobson, and Sanja Fidler.
\newblock {Learning to predict 3D objects with an interpolation-based
  differentiable renderer}.
\newblock {\em NeurIPS}, 2019.

\bibitem{chen2020neural}
Zhang Chen, Anpei Chen, Guli Zhang, Chengyuan Wang, Yu Ji, Kiriakos~N
  Kutulakos, and Jingyi Yu.
\newblock A neural rendering framework for free-viewpoint relighting.
\newblock In {\em CVPR}, 2020.

\bibitem{cook1982reflectance}
Robert~L Cook and Kenneth~E. Torrance.
\newblock A reflectance model for computer graphics.
\newblock {\em ACM TOG}, 1(1):7--24, 1982.

\bibitem{debevec2000acquiring}
Paul Debevec, Tim Hawkins, Chris Tchou, Haarm-Pieter Duiker, Westley Sarokin,
  and Mark Sagar.
\newblock Acquiring the reflectance field of a human face.
\newblock In {\em ACM SIGGRAPH}, 2000.

\bibitem{egger20203dmm}
Bernhard Egger, William~AP Smith, Ayush Tewari, Stefanie Wuhrer, Michael
  Zollhoefer, Thabo Beeler, Florian Bernard, Timo Bolkart, Adam Kortylewski,
  Sami Romdhani, Christian Theobalt, Volker Blanz, and Thomas Vetter.
\newblock {3D morphable face models—past, present, and future}.
\newblock {\em ACM TOG}, 39(5):1--38, 2020.

\bibitem{fridovich2022plenoxels}
Sara Fridovich-Keil, Alex Yu, Matthew Tancik, Qinhong Chen, Benjamin Recht, and
  Angjoo Kanazawa.
\newblock {Plenoxels: Radiance fields without neural networks}.
\newblock In {\em CVPR}, 2022.

\bibitem{ghosh2011multiview}
Abhijeet Ghosh, Graham Fyffe, Borom Tunwattanapong, Jay Busch, Xueming Yu, and
  Paul Debevec.
\newblock Multiview face capture using polarized spherical gradient
  illumination.
\newblock In {\em ACM SIGGRAPH Asia}, 2011.

\bibitem{ghosh2008practical}
Abhijeet Ghosh, Tim Hawkins, Pieter Peers, Sune Frederiksen, and Paul Debevec.
\newblock Practical modeling and acquisition of layered facial reflectance.
\newblock In {\em ACM SIGGRAPH Asia}. 2008.

\bibitem{gotardo2018practical}
Paulo Gotardo, J{\'e}r{\'e}my Riviere, Derek Bradley, Abhijeet Ghosh, and Thabo
  Beeler.
\newblock Practical dynamic facial appearance modeling and acquisition.
\newblock {\em ACM TOG}, 37(6):1--13, 2018.

\bibitem{hasselgren2022shape}
Jon Hasselgren, Nikolai Hofmann, and Jacob Munkberg.
\newblock {Shape, light \& material decomposition from images using Monte Carlo
  rendering and denoising}.
\newblock In {\em NeurIPS}, 2022.

\bibitem{igarashi2005appearance}
Takanori Igarashi, Ko Nishino, and Shree~K Nayar.
\newblock The appearance of human skin.
\newblock 2005.

\bibitem{jensen2001practical}
Henrik~Wann Jensen, Stephen~R Marschner, Marc Levoy, and Pat Hanrahan.
\newblock A practical model for subsurface light transport.
\newblock In {\em ACM SIGGRAPH}, 2001.

\bibitem{jensen2014large}
Rasmus Jensen, Anders Dahl, George Vogiatzis, Engin Tola, and Henrik Aan{\ae}s.
\newblock Large scale multi-view stereopsis evaluation.
\newblock In {\em CVPR}, 2014.

\bibitem{kajiya1986rendering}
James~T Kajiya.
\newblock The rendering equation.
\newblock In {\em ACM SIGGRAPH}, 1986.

\bibitem{kampouris2018diffuse}
Christos Kampouris, Stefanos Zafeiriou, and Abhijeet Ghosh.
\newblock Diffuse-specular separation using binary spherical gradient
  illumination.
\newblock In {\em EGSR}, 2018.

\bibitem{karis2013real}
Brian Karis and Epic Games.
\newblock {Real shading in Unreal Engine 4}.
\newblock {\em Proc. Physically Based Shading Theory Practice}, 4(3):1, 2013.

\bibitem{klehm2015recent}
Oliver Klehm, Fabrice Rousselle, Marios Papas, Derek Bradley, Christophe Hery,
  Bernd Bickel, Wojciech Jarosz, and Thabo Beeler.
\newblock Recent advances in facial appearance capture.
\newblock In {\em CGF}, 2015.

\bibitem{li2022eyenerf}
Gengyan Li, Abhimitra Meka, Franziska Mueller, Marcel~C Buehler, Otmar
  Hilliges, and Thabo Beeler.
\newblock Eyenerf: a hybrid representation for photorealistic synthesis,
  animation and relighting of human eyes.
\newblock {\em ACM TOG}, 41(4):1--16, 2022.

\bibitem{liu2020dist}
Shaohui Liu, Yinda Zhang, Songyou Peng, Boxin Shi, Marc Pollefeys, and Zhaopeng
  Cui.
\newblock {DIST: Rendering deep implicit signed distance function with
  differentiable sphere tracing}.
\newblock In {\em CVPR}, 2020.

\bibitem{lyu2022neural}
Linjie Lyu, Ayush Tewari, Thomas Leimk{\"u}hler, Marc Habermann, and Christian
  Theobalt.
\newblock Neural radiance transfer fields for relightable novel-view synthesis
  with global illumination.
\newblock In {\em ECCV}, 2022.

\bibitem{ma2007rapid}
Wan-Chun Ma, Tim Hawkins, Pieter Peers, Charles-Felix Chabert, Malte Weiss, and
  Paul~E Debevec.
\newblock Rapid acquisition of specular and diffuse normal maps from polarized
  spherical gradient illumination.
\newblock {\em ECRT}, 2007(9):183--194, 2007.

\bibitem{martin2021nerfw}
Ricardo Martin-Brualla, Noha Radwan, Mehdi~SM Sajjadi, Jonathan~T Barron,
  Alexey Dosovitskiy, and Daniel Duckworth.
\newblock {NeRF in the Wild: Neural radiance fields for unconstrained photo
  collections}.
\newblock In {\em CVPR}, 2021.

\bibitem{merl}
Wojciech Matusik, Hanspeter Pfister, Matt Brand, and Leonard McMillan.
\newblock A data-driven reflectance model.
\newblock {\em ACM TOG}, 22(3):759--769, 2003.

\bibitem{mildenhall2022nerfd}
Ben Mildenhall, Peter Hedman, Ricardo Martin-Brualla, Pratul~P Srinivasan, and
  Jonathan~T Barron.
\newblock {NeRF in the Dark: High dynamic range view synthesis from noisy raw
  images}.
\newblock In {\em CVPR}, 2022.

\bibitem{mildenhall2020nerf}
B Mildenhall, PP Srinivasan, M Tancik, JT Barron, R Ramamoorthi, and R Ng.
\newblock {NeRF: Representing scenes as neural radiance fields for view
  synthesis}.
\newblock In {\em ECCV}, 2020.

\bibitem{mueller2022instant}
Thomas M\"uller, Alex Evans, Christoph Schied, and Alexander Keller.
\newblock Instant neural graphics primitives with a multiresolution hash
  encoding.
\newblock {\em ACM TOG}, 41(4):102:1--102:15, 2022.

\bibitem{munkberg2022extracting}
Jacob Munkberg, Jon Hasselgren, Tianchang Shen, Jun Gao, Wenzheng Chen, Alex
  Evans, Thomas M{\"u}ller, and Sanja Fidler.
\newblock {Extracting triangular 3D models, materials, and lighting from
  images}.
\newblock In {\em CVPR}, 2022.

\bibitem{murakami2002measurement}
H Murakami, T Horii, N Tsumura, and Y Miyake.
\newblock {Measurement and simulation of 3D gonio spectral reflectance of skin
  surface}.
\newblock {\em DBJ}, 93:21--26, 2002.

\bibitem{oechsle2021unisurf}
Michael Oechsle, Songyou Peng, and Andreas Geiger.
\newblock {UNISURF: Unifying neural implicit surfaces and radiance fields for
  multi-view reconstruction}.
\newblock In {\em CVPR}, 2021.

\bibitem{pharr2016physically}
Matt Pharr, Wenzel Jakob, and Greg Humphreys.
\newblock {\em {Physically based rendering: From theory to implementation}}.
\newblock Morgan Kaufmann, 2016.

\bibitem{phong1975illumination}
Bui~Tuong Phong.
\newblock Illumination for computer generated pictures.
\newblock {\em Communications of the ACM}, 18(6):311--317, 1975.

\bibitem{pumarola2021dnerf}
Albert Pumarola, Enric Corona, Gerard Pons-Moll, and Francesc Moreno-Noguer.
\newblock {D-NeRF: Neural radiance fields for dynamic scenes}.
\newblock In {\em CVPR}, 2021.

\bibitem{ramamoorthi2002frequency}
Ravi Ramamoorthi and Pat Hanrahan.
\newblock Frequency space environment map rendering.
\newblock In {\em ACM SIGGRAPH}, 2002.

\bibitem{reiser2021kilonerf}
Christian Reiser, Songyou Peng, Yiyi Liao, and Andreas Geiger.
\newblock {KiloNeRF: Speeding up neural radiance fields with thousands of tiny
  MLPs}.
\newblock In {\em CVPR}, 2021.

\bibitem{riviere2020single}
J{\'e}r{\'e}my Riviere, Paulo Gotardo, Derek Bradley, Abhijeet Ghosh, and Thabo
  Beeler.
\newblock Single-shot high-quality facial geometry and skin appearance capture.
\newblock {\em ACM TOG}, 39(4):81:1--81:12, 2020.

\bibitem{sitzmann2019siren}
Vincent Sitzmann, Julien~N.P. Martel, Alexander~W. Bergman, David~B. Lindell,
  and Gordon Wetzstein.
\newblock Implicit neural representations with periodic activation functions.
\newblock In {\em NeurIPS}, 2020.

\bibitem{srinivasan2021nerv}
Pratul~P Srinivasan, Boyang Deng, Xiuming Zhang, Matthew Tancik, Ben
  Mildenhall, and Jonathan~T Barron.
\newblock {NeRV: Neural reflectance and visibility fields for relighting and
  view synthesis}.
\newblock In {\em CVPR}, 2021.

\bibitem{sun2021nelf}
Tiancheng Sun, Kai-En Lin, Sai Bi, Zexiang Xu, and Ravi Ramamoorthi.
\newblock Nelf: Neural light-transport field for portrait view synthesis and
  relighting.
\newblock {\em EGSR}, 2021.

\bibitem{nvidia2021realistic}
Blago Taskov, Eugene d'Eon, Morteza Ramezanali, and Simon Yuen.
\newblock {Realistic digital human rendering with Omniverse RTX Renderer}.
\newblock
  \url{https://www.nvidia.com/en-us/on-demand/session/siggraph2021-sigg21-s-09/}.

\bibitem{tewari2022advances}
Ayush Tewari, Justus Thies, Ben Mildenhall, Pratul Srinivasan, Edgar Tretschk,
  Wang Yifan, Christoph Lassner, Vincent Sitzmann, Ricardo Martin-Brualla,
  Stephen Lombardi, Tomas Simon, Christian Theobalt, Matthias Niessner,
  Jonathan~T Barron, Gordon Wetzstein, Michael Zollhoefer, and Vladislav
  Golyanik.
\newblock Advances in neural rendering.
\newblock In {\em CGF}, 2022.

\bibitem{torrance1967theory}
Kenneth~E Torrance and Ephraim~M Sparrow.
\newblock Theory for off-specular reflection from roughened surfaces.
\newblock {\em JOSA}, 57(9):1105--1114, 1967.

\bibitem{tunwattanapong2013acquiring}
Borom Tunwattanapong, Graham Fyffe, Paul Graham, Jay Busch, Xueming Yu,
  Abhijeet Ghosh, and Paul Debevec.
\newblock Acquiring reflectance and shape from continuous spherical harmonic
  illumination.
\newblock {\em ACM TOG}, 32(4):1--12, 2013.

\bibitem{valery2000tissue}
Tuchin Valery.
\newblock {Tissue optics: Light scattering methods and instruments for medical
  diagnosis}.
\newblock {\em SPIE Tutorial Texts in Optical Engineering}, 2000.

\bibitem{verbin2022ref}
Dor Verbin, Peter Hedman, Ben Mildenhall, Todd Zickler, Jonathan~T Barron, and
  Pratul~P Srinivasan.
\newblock {Ref-NeRF: Structured view-dependent appearance for neural radiance
  fields}.
\newblock In {\em CVPR}, 2022.

\bibitem{wang2021neus}
Peng Wang, Lingjie Liu, Yuan Liu, Christian Theobalt, Taku Komura, and Wenping
  Wang.
\newblock {NeuS: Learning neural implicit surfaces by volume rendering for
  multi-view reconstruction}.
\newblock {\em NeurIPS}, 2021.

\bibitem{wang2022hfneus}
Yiqun Wang, Ivan Skorokhodov, and Peter Wonka.
\newblock {HF-NeuS: Improved surface reconstruction using high-frequency
  details}.
\newblock In {\em NeurIPS}, 2022.

\bibitem{wang2004image}
Zhou Wang, Alan~C Bovik, Hamid~R Sheikh, and Eero~P Simoncelli.
\newblock {Image quality assessment: From error visibility to structural
  similarity}.
\newblock {\em IEEE TIP}, 13(4):600--612, 2004.

\bibitem{weyrich2006analysis}
Tim Weyrich, Wojciech Matusik, Hanspeter Pfister, Bernd Bickel, Craig Donner,
  Chien Tu, Janet McAndless, Jinho Lee, Addy Ngan, Henrik~Wann Jensen, and
  Markus Gross.
\newblock Analysis of human faces using a measurement-based skin reflectance
  model.
\newblock {\em ACM TOG}, 25(3):1013--1024, 2006.

\bibitem{yang2020facescape}
Haotian Yang, Hao Zhu, Yanru Wang, Mingkai Huang, Qiu Shen, Ruigang Yang, and
  Xun Cao.
\newblock {Facescape: A large-scale high quality 3D face dataset and detailed
  riggable 3D face prediction}.
\newblock In {\em CVPR}, 2020.

\bibitem{yao2022neilf}
Yao Yao, Jingyang Zhang, Jingbo Liu, Yihang Qu, Tian Fang, David McKinnon,
  Yanghai Tsin, and Long Quan.
\newblock {NeILF: Neural incident light field for physically-based material
  estimation}.
\newblock In {\em ECCV}, 2022.

\bibitem{yariv2021volsdf}
Lior Yariv, Jiatao Gu, Yoni Kasten, and Yaron Lipman.
\newblock Volume rendering of neural implicit surfaces.
\newblock {\em NeurIPS}, 2021.

\bibitem{yariv2020idr}
Lior Yariv, Yoni Kasten, Dror Moran, Meirav Galun, Matan Atzmon, Basri Ronen,
  and Yaron Lipman.
\newblock Multiview neural surface reconstruction by disentangling geometry and
  appearance.
\newblock {\em NeurIPS}, 2020.

\bibitem{zhang2021physg}
Kai Zhang, Fujun Luan, Qianqian Wang, Kavita Bala, and Noah Snavely.
\newblock {PhySG: Inverse rendering with spherical gaussians for physics-based
  material editing and relighting}.
\newblock In {\em CVPR}, 2021.

\bibitem{zhang2020nerfpp}
Kai Zhang, Gernot Riegler, Noah Snavely, and Vladlen Koltun.
\newblock {NeRF++: Analyzing and improving neural radiance fields}.
\newblock {\em arXiv preprint arXiv:2010.07492}, 2020.

\bibitem{zhang2018unreasonable}
Richard Zhang, Phillip Isola, Alexei~A Efros, Eli Shechtman, and Oliver Wang.
\newblock The unreasonable effectiveness of deep features as a perceptual
  metric.
\newblock In {\em CVPR}, 2018.

\bibitem{zhang2021nerfactor}
Xiuming Zhang, Pratul~P Srinivasan, Boyang Deng, Paul Debevec, William~T
  Freeman, and Jonathan~T Barron.
\newblock {NeRFactor: Neural factorization of shape and reflectance under an
  unknown illumination}.
\newblock {\em ACM TOG}, 40(6):1--18, 2021.

\bibitem{zheng2021compact}
Chuankun Zheng, Ruzhang Zheng, Rui Wang, Shuang Zhao, and Hujun Bao.
\newblock {A compact representation of measured BRDFs using neural processes}.
\newblock {\em ACM TOG}, 41(2):1--15, 2021.

\bibitem{zheng2022imface}
Mingwu Zheng, Hongyu Yang, Di Huang, and Liming Chen.
\newblock {ImFace: A nonlinear 3D morphable face model with implicit neural
  representations}.
\newblock In {\em CVPR}, 2022.

\end{thebibliography}
}

\clearpage
\appendix  
\setcounter{table}{0}  
\setcounter{figure}{0}
\renewcommand\thesection{\Alph{section}}
\renewcommand{\thetable}{A\arabic{table}}
\renewcommand{\thefigure}{A\arabic{figure}}

\begin{strip}%
  \centering
  \large
  \textbf{%
  {NeuFace: Realistic 3D Neural Face Rendering from Multi-view Images}\\
  \vspace{0.3cm} \textit{Supplementary material} \\ \vspace{0.5cm}
  }
  \includegraphics[width=1\linewidth]{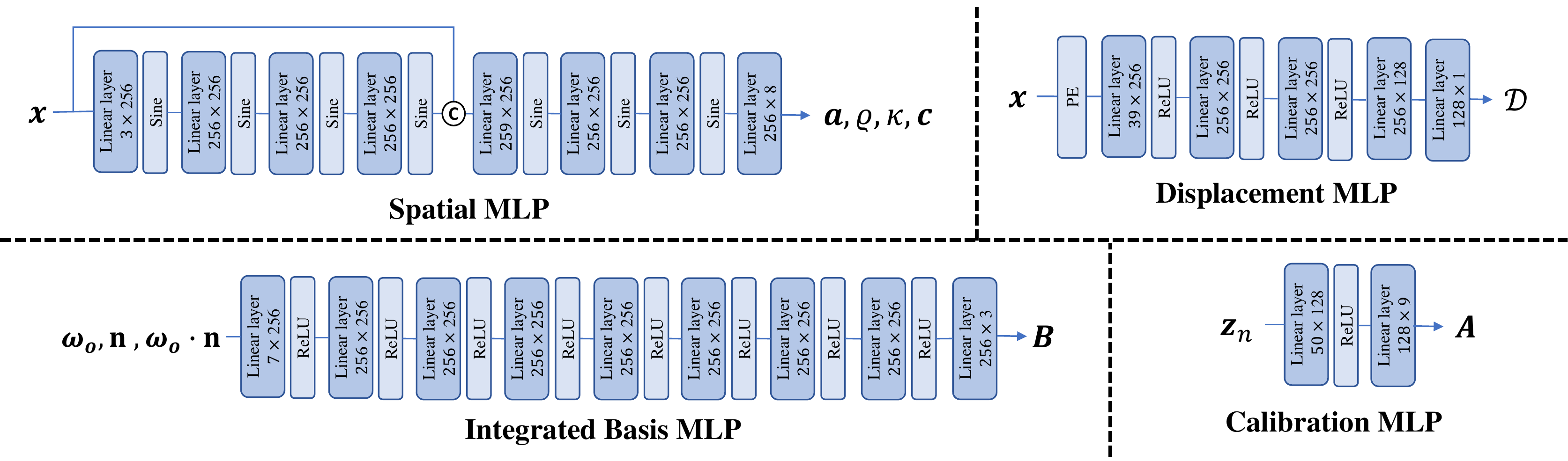}
  \captionof{figure}{Detailed architecture of the NeuFace components.}
  \label{fig:supp_net}
\end{strip}


\section{Light Integral with vMF distribution}
\label{sec:light}

\begin{figure}
  \centering
  \setlength{\abovecaptionskip}{0pt}
  \setlength{\belowcaptionskip}{0pt}
  \includegraphics[width=0.9\linewidth]{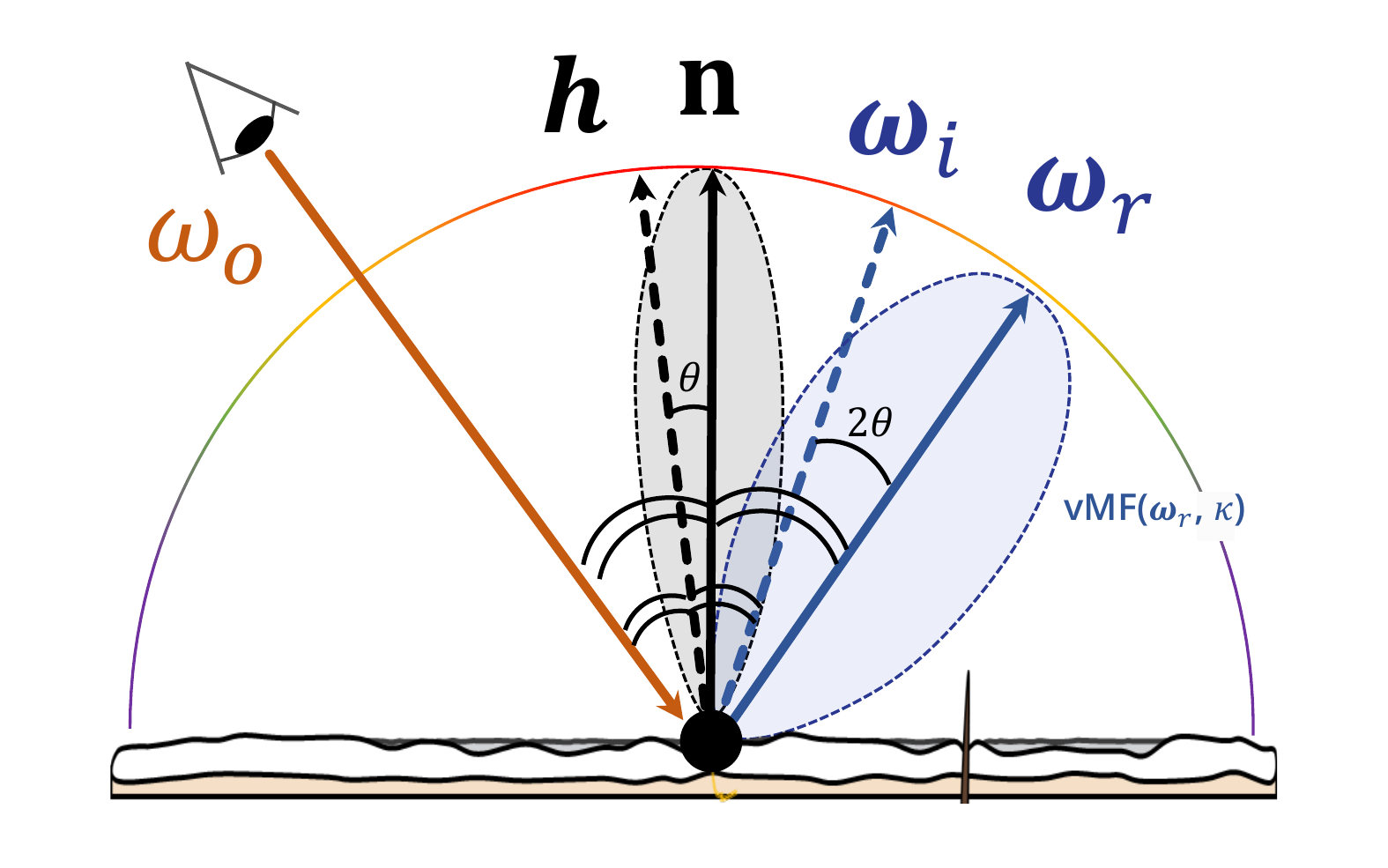}
  \caption{Illustration of the direction vectors. The gray lobe indicates the distribution of half vector $h$. In our case, the distribution is symmetric about the normal vector, similar to the case in microfacet models.}
  \label{fig:supp_prof}
\end{figure}

We provide the proof for the light integral approximation in our model,
\textit{i.e.} Eq.~\eqref{spec_light} in the main paper. Considering that the reflected energy is mostly concentrated near the reflection direction, a 3-dimensional normalized Spherical Gaussian distribution (von Mises–Fisher distribution, $\operatorname{vMF}$) is assumed:
\begin{equation}
     D(h) \approx \frac{\kappa}{4\pi\sinh{\kappa}} e^{\kappa {\omega_{r}^{\top}} {\omega_i}},    
\end{equation}
\noindent where $h$ indicates the half vector, which is a function of random variable ${\omega_i}$. We use $D(h)$ instead of $D(h(\omega_i,\omega_{o});\kappa)$ for abbreviation. $D(h)$ is symmetric about the normal vector, similar to the case in microfacet models, as shown in Fig.~\ref{fig:supp_prof}.

Based on Eq.~\eqref{sh_l} in our main paper, the light integral can be expressed as an expected value:
\begin{equation}
\begin{aligned}
     & \int_{\mathrm{s}^2} D(h) L_i\left(\omega_i\right) \mathrm{d} \omega_i 
      = \mathbb{E}_{\omega_i \sim \operatorname{vMF}(\omega_r,\kappa)}[ L_i\left(\omega_i\right)] \\
      & = \mathbb{E}_{\omega_i \sim \operatorname{vMF}(\omega_r,\kappa)}[ \sum_{\ell=0} \sum_{m=-\ell}^\ell c_{\ell m}  {\mit Y}_{\ell m}(\omega_i)] \\
      & = \sum_{\ell=0} \sum_{m=-\ell}^\ell c_{\ell m} \mathbb{E}_{\omega_i \sim \operatorname{vMF}(\omega_r,\kappa)}[{\mit Y}_{\ell m}(\omega_i)]. 
      \label{eq:expected}
\end{aligned}
\end{equation}
\hspace{1em} In \cite{verbin2022ref}, it is proved that the expected value of a spherical harmonic function under a vMF distribution has the approximation as follows:
\vspace{-2mm}
\begin{equation}
\begin{aligned}
    \mathbb{E}_{\omega_i \sim \operatorname{vMF}(\omega_r,\kappa)}[{\mit Y}_{\ell m}(\omega_i)]
    \approx e^{-\frac{\ell(\ell+1)}{2\kappa}} {\mit Y}_{\ell m}(\omega_r),
     \label{eq:approximation}
\end{aligned}
\end{equation}
with $O(\kappa^{-2})$ as an error bound. 

By substituting Eq.~\eqref{eq:approximation} into \eqref{eq:expected}, we have:
{\small
\begin{equation}
      \int_{\mathrm{s}^2} D(h) L_i\left(\omega_i\right) \mathrm{d} \omega_i 
      \approx\sum_{\ell=0} \sum_{m=-\ell}^\ell e^{-\frac{\ell(\ell+1)}{2\kappa}} c_{\ell m} {\mit Y}_{\ell m}(\omega_r).
\end{equation}}

\begin{table*}[t]
    \resizebox{1.\textwidth}{!}{
    \begin{tabular}{l|cccc|cccc|cccc|cccc}
    & \multicolumn{4}{c|}{\textbf{Sub.~2 Exp.~4}}
    & \multicolumn{4}{c|}{\textbf{Sub.~2 Exp.~16}}
    & \multicolumn{4}{c|}{\textbf{Sub.~3 Exp.~4}}
    & \multicolumn{4}{c}{\textbf{Sub.~3 Exp.~16}} \\
    & \multicolumn{1}{c}{\scriptsize PSNR $\uparrow$} 
    & \multicolumn{1}{c}{\scriptsize SSIM $\uparrow$}
    & \multicolumn{1}{c}{\scriptsize LPIPS $\downarrow$}
    & \multicolumn{1}{c|}{\scriptsize Chamfer $\downarrow$} 
    & \multicolumn{1}{c}{\scriptsize PSNR $\uparrow$} 
    & \multicolumn{1}{c}{\scriptsize SSIM $\uparrow$}
    & \multicolumn{1}{c}{\scriptsize LPIPS $\downarrow$}
    & \multicolumn{1}{c|}{\scriptsize Chamfer $\downarrow$} 
    & \multicolumn{1}{c}{\scriptsize PSNR $\uparrow$} 
    & \multicolumn{1}{c}{\scriptsize SSIM $\uparrow$}
    & \multicolumn{1}{c}{\scriptsize LPIPS $\downarrow$}
    & \multicolumn{1}{c|}{\scriptsize Chamfer $\downarrow$} 
    & \multicolumn{1}{c}{\scriptsize PSNR $\uparrow$} 
    & \multicolumn{1}{c}{\scriptsize SSIM $\uparrow$}
    & \multicolumn{1}{c}{\scriptsize LPIPS $\downarrow$} 
    & \multicolumn{1}{c}{\scriptsize Chamfer $\downarrow$} \\
    \shline
     NeRF\cite{mildenhall2020nerf} 
     & \gc{12.35} & \gc{0.671} & \gc{.1433} & \gc{3.311} 
     & \gc{-} & \gc{-} & \gc{-} & \gc{-}
     & \gc{-} & \gc{-} & \gc{-} & \gc{-}
     & \gc{-} & \gc{-} & \gc{-} & \gc{-}   \\
     Ref-NeRF\cite{verbin2022ref} 
     & {19.26} & {0.833} & {.1231} & {3.219}
     & {18.51} & {0.800} & {.1311} & {3.303}
     & {17.16} & {0.801} & {.1452} & {5.862}
     & {17.93} & {0.811} & {.1294} & {3.725}\\
     VolSDF\cite{yariv2021volsdf} 
     & \cellcolor{orange}{29.06} & \cellcolor{orange}{0.937} & \cellcolor{lightyellow}{.0598} & \cellcolor{orange}{0.508}
     & \cellcolor{lightyellow}{26.95} & \cellcolor{lightyellow}{0.927} & \cellcolor{lightyellow}{.0618} & \cellcolor{orange}{0.514}
     & \cellcolor{orange}{29.45} & \cellcolor{orange}{0.939} & \cellcolor{lightyellow}{.0540} & \cellcolor{orange}{0.637}
     & \cellcolor{orange}{28.00} & \cellcolor{orange}{0.929} & \cellcolor{lightyellow}{.0540} & \cellcolor{orange}{0.578}\\
     PhySG\cite{zhang2021physg} 
     & {25.56} & {0.918} & {.0676} & {0.972}
     & {24.12} & {0.905} & {.0704} & {1.770}
     & {23.67} & {0.892} & {.0770} & {1.336}
     & {23.08} & {0.877} & {.0797} & {2.031}\\
     DIFFREC\cite{munkberg2022extracting} 
     & \cellcolor{lightyellow}{27.52} & \cellcolor{lightyellow}{0.932} & \cellcolor{orange}{.0254} & \cellcolor{lightyellow}{0.612}
     & \cellcolor{orange}{27.04} & \cellcolor{orange}{0.930} & \cellcolor{orange}{.0271} & \cellcolor{lightyellow}{0.621}
     & \cellcolor{lightyellow}{26.72} & \cellcolor{lightyellow}{0.928} & \cellcolor{orange}{.0292} & \cellcolor{lightyellow}{0.706}
     & \cellcolor{lightyellow}{26.35} & \cellcolor{lightyellow}{0.923} & \cellcolor{orange}{.0295} & \cellcolor{lightyellow}{0.889} \\
     \hline
     \textit{NeuFace}  
     & \cellcolor{tablered}\textbf{30.14} &  \cellcolor{tablered}\textbf{0.954} &  \cellcolor{tablered}\textbf{.0235} & \cellcolor{tablered}\textbf{0.439}
     & \cellcolor{tablered}\textbf{28.77} &  \cellcolor{tablered}\textbf{0.946} &  \cellcolor{tablered}\textbf{.0256} & \cellcolor{tablered}\textbf{0.467}
     &  \cellcolor{tablered}\textbf{30.86} &  \cellcolor{tablered}\textbf{0.956} &  \cellcolor{tablered}\textbf{.0268} & \cellcolor{tablered}\textbf{0.535}
     &  \cellcolor{tablered}\textbf{30.13} &  \cellcolor{tablered}\textbf{0.949} &  \cellcolor{tablered}\textbf{.0247} & \cellcolor{tablered}\textbf{0.528}\\
    \end{tabular}}
    \caption{PSNRs, SSIMs, LPIPSs, and Chamfer distances for the per-scene test set.
    \label{tab:supp_result}
    }
\vspace{-2mm}
\end{table*}

\section{Implementation Details}
 \para{Network Architecture} All the networks in \textit{NeuFace} are fully implemented by MLPs, and the detailed architectures of the Spatial MLP, the displacement MLP, the Integrated Basis MLP, and the calibration MLP are individually shown in Fig.~\ref{fig:supp_net}. For improved performance on high-frequency clues, in the displacement MLP, we encode the coordinates by sinusoidal positional encoding $\gamma$\cite{mildenhall2020nerf}, written as $\gamma(x)\!=\!(\sin(2^0\pi x),\cos(2^0\pi x),\cdots,\sin(2^{L-1}\pi x),\cos(2^{L-1}\pi x))$, where $L\!=\!6$ in our experiments. Besides, sinusoidal activations are exploited in the Spatial MLP and the parameters are initialized as in \cite{sitzmann2019siren}, where the albedo ${\bf a}$ and the specular intensity $\varrho $ are output using a sigmoid activation and the shininess $\kappa$ is predicted as an inverse $1/\kappa$ using a softplus function as \cite{verbin2022ref} does. Also, an ReLU activation is applied to the diffuse (Eq.~\eqref{diff_final}) and specular light transport (Eq.~\eqref{spec_light}) to avoid negative light values. 
 
\begin{figure}
  \centering
  \setlength{\abovecaptionskip}{0pt}
  \setlength{\belowcaptionskip}{0pt}
  \includegraphics[width=1.\linewidth]{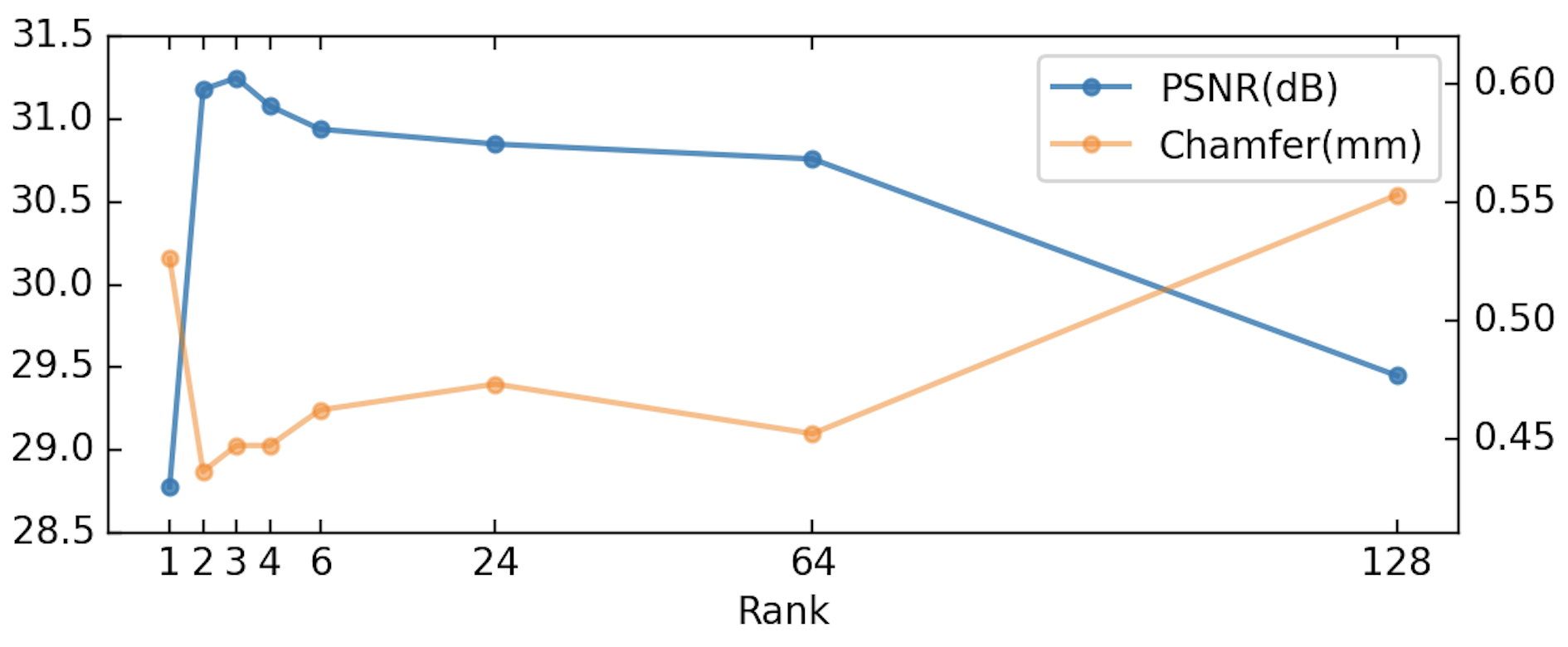}
   \caption{The PSNR and Chamfer distance curves over different rank values.
   }
   \label{fig:rank}
\end{figure}

\begin{figure}
  \centering
  \setlength{\abovecaptionskip}{0pt}
  \setlength{\belowcaptionskip}{0pt}
  \includegraphics[width=1.\linewidth]{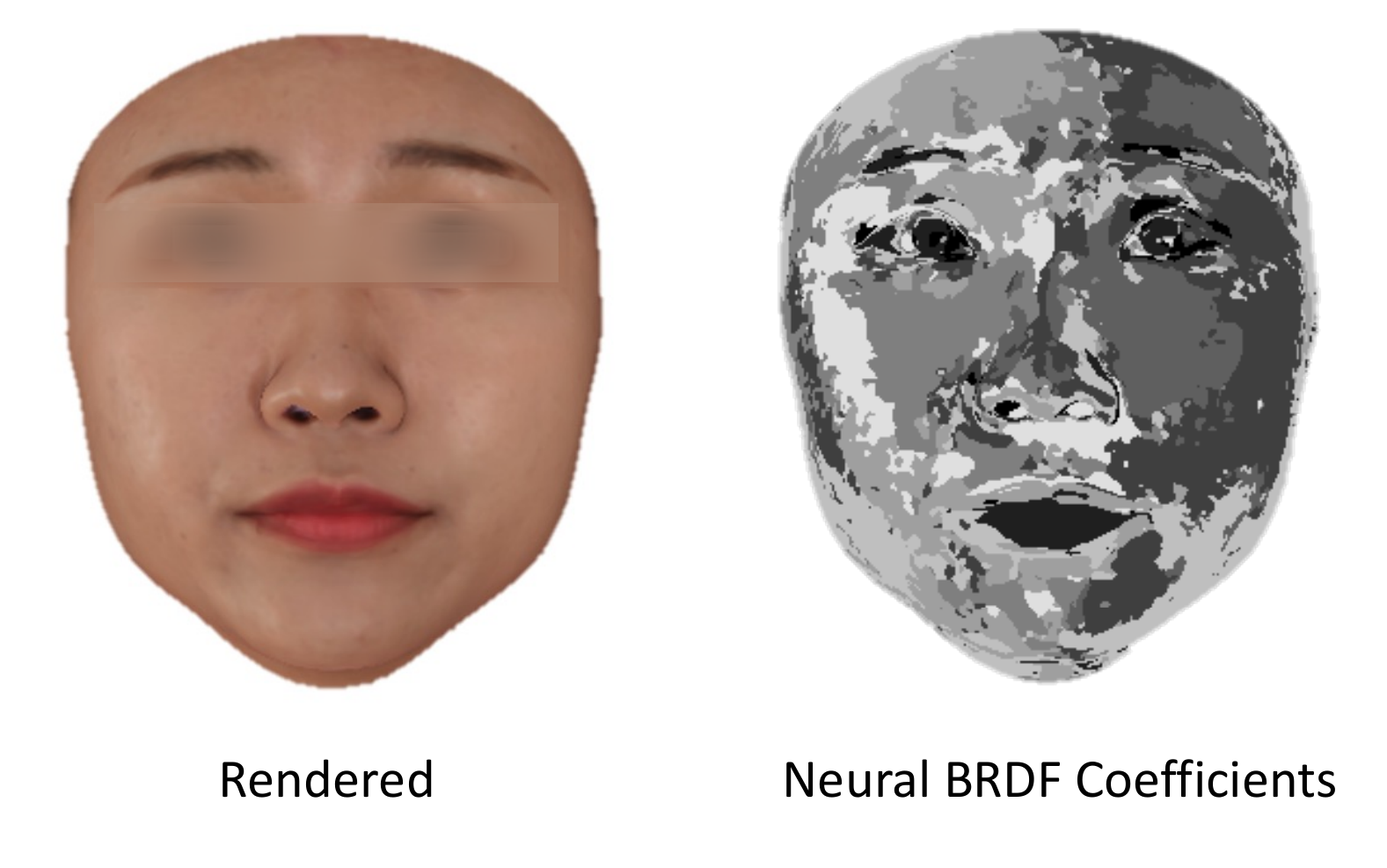}
   \caption{Demonstration of the rendered image and clustered neural BRDFs coefficients.}
   \label{fig:c_kmeans}
\end{figure}

\begin{table}
    \centering
    \resizebox{.48\textwidth}{!}{
    \begin{tabular}{@{}l|cccc}
    Designs & PSNR $\uparrow$ & SSIM $\uparrow$ & LPIPS $\downarrow$ & Chamfer $\downarrow$ \\
    \shline
    w/o geometry prior & \cellcolor{lightyellow}{28.11} & \cellcolor{lightyellow}0.942 & \cellcolor{lightyellow}.0452  & \cellcolor{lightyellow}0.647 \\
    w/o specular loss & \cellcolor{orange}{31.03} & \cellcolor{orange}0.957 & \cellcolor{orange}.0248 & \cellcolor{orange}0.453 \\
    \hline
    \textit{NeuFace} &  \cellcolor{tablered}\textbf{31.25} &  \cellcolor{tablered}\textbf{0.958} &   \cellcolor{tablered}\textbf{.0237} &  \cellcolor{tablered}\textbf{0.447} \\
    \end{tabular}} 
    \caption{Quantitative ablation study on neural photometric calibration, geometry prior, and the specular energy loss.}
    \label{table:miscellaneous_ablation}
\end{table}

\para{Training Details} The model is trained with an Adam optimizer in an end-to-end manner for 3,000 epochs in total with the initial learning rate set at 0.0001 except for \textit{ImFace} ~\cite{zheng2022imface}, which is fine-tuned using 1\% of the learning rate. The learning rate decays by a factor of 0.5 for every 375 epochs until 75\% of the training process. It takes around 15 hours to train the model on 4 NVIDIA RTX 3080ti GPUs with a mini-batch of 2,048 rays. The trade-off parameters $\lambda_1, \lambda_2, \lambda_4,  \lambda_5,\lambda_6$ are set to $1$, $5\!\times\!10^{-3}$, $10^{3}$, $1\!\times\!10^{-1}$, and $10^{-3}$, respectively. $\lambda_3=8\times 10^{-3}$ achieves decent performance and is slightly adjusted for different individuals.

\para{Testing Details} We remove the background of all images using the ground-truth 3D model and the \textit{ImFace} boundary. To align the photometric properties between the rendered and test images, the optimal linear mapping matrixes $\mathbf{A}$ is calculated before evaluation by pseudo-inverse of RGB values. In the relighting evaluation, HDR environment lights are transformed to Spherical Harmonics as in \cite{chen2020neural}. It takes about 2 minutes to render a 1K image on a single NVIDIA RTX 3080ti GPU during testing.

\para{Volume Rendering Details} We perform aggressive sphere tracing to accelerate the rendering procedure. To avoid tracing across the facial surface at the first step, the SDF value is truncated to $[-5cm,5cm]$. During the tracing process, we update the ray with 1.2 times the queried signed distance value \cite{liu2020dist}. Since the geometric prior \textit{ImFace} defines the facial surface within a unit sphere, where the SDF values in the backspace are negative, we assume the rays which enter the sphere from the negative positions have no intersection with the surface (denoted as `unhit'). 

\begin{figure*}
  \setlength{\abovecaptionskip}{0pt}
  \setlength{\belowcaptionskip}{0pt}
  \includegraphics[width=1.0\linewidth]{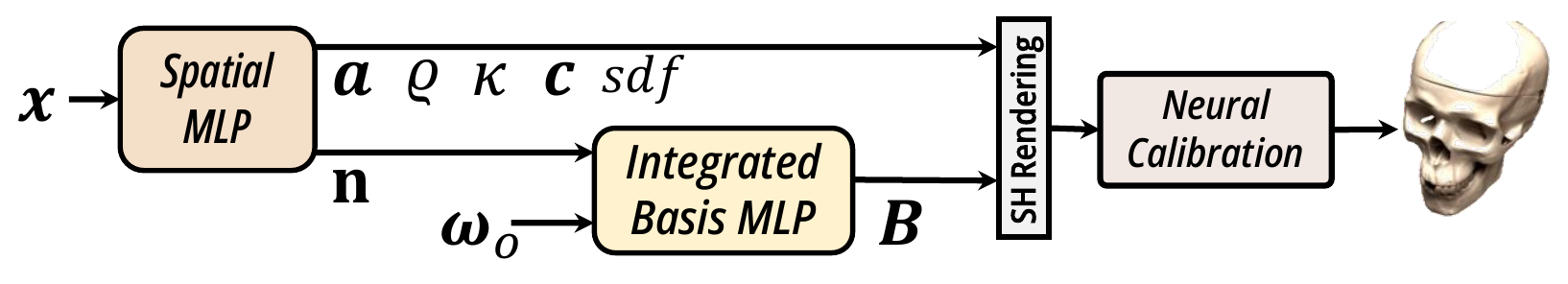}
  \caption{Network architecture on general objects.}
  \label{fig:dtu_pipeline}
\end{figure*}

\begin{figure*}
  \setlength{\abovecaptionskip}{0pt}
  \setlength{\belowcaptionskip}{0pt}
  \includegraphics[width=1.0\linewidth]{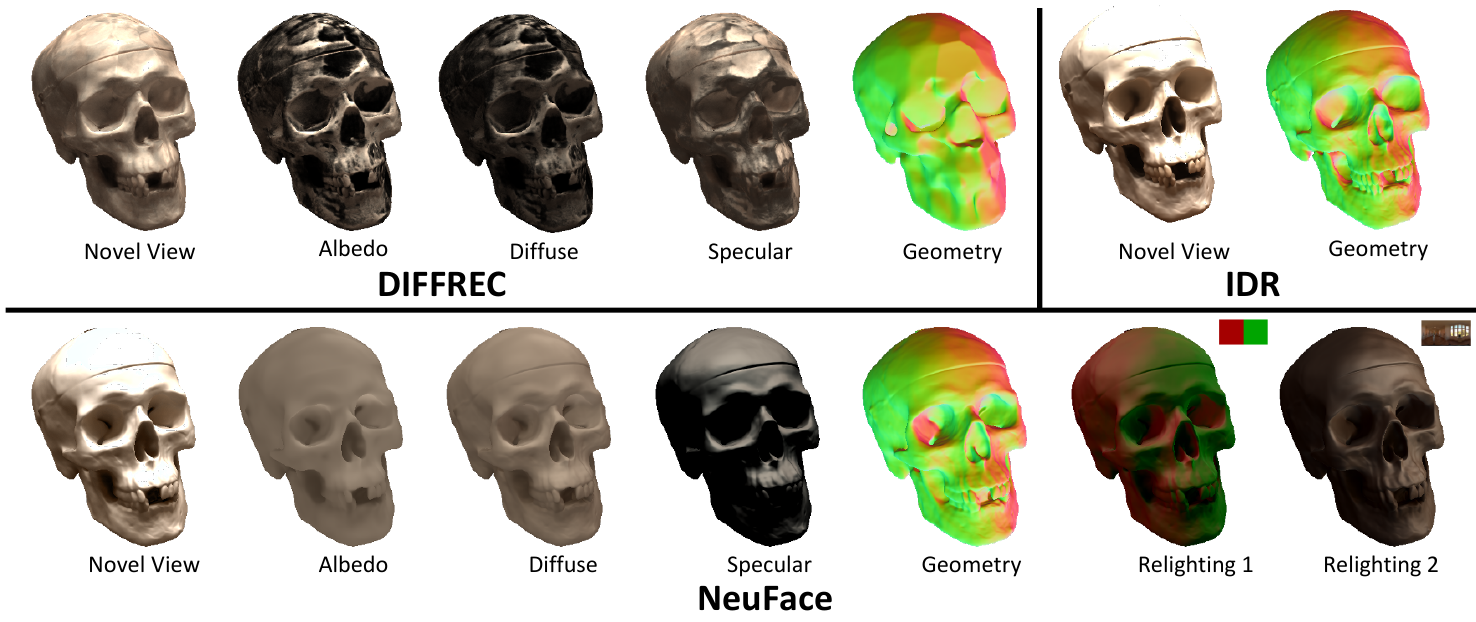}
  \caption{Reconstruction results of scan 65 on the DTU dataset\cite{jensen2014large} (zoom in for a better view).}
  \label{fig:dtu65}
\end{figure*}

\begin{figure*}
  \setlength{\abovecaptionskip}{0pt}
  \setlength{\belowcaptionskip}{0pt}
  \includegraphics[width=1.0\linewidth]{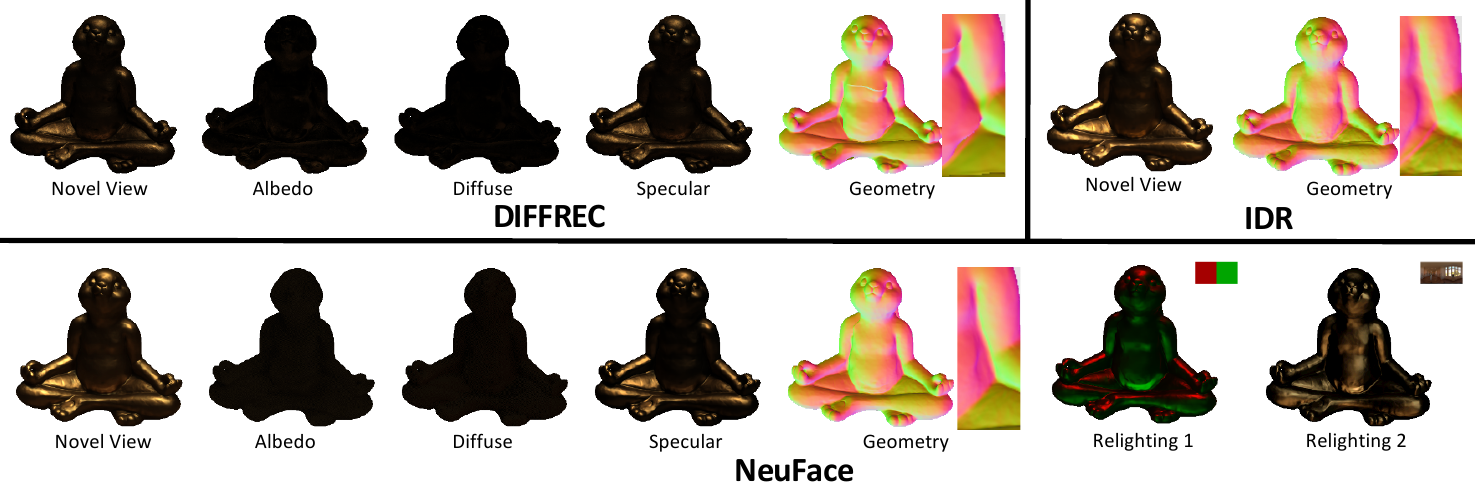}
  \caption{Reconstruction results of scan 110 on the DTU dataset\cite{jensen2014large} (zoom in for a better view).}
  \label{fig:dtu110}
\end{figure*}

\begin{figure*}
  \setlength{\abovecaptionskip}{0pt}
  \setlength{\belowcaptionskip}{0pt}
  \includegraphics[width=1.0\linewidth]{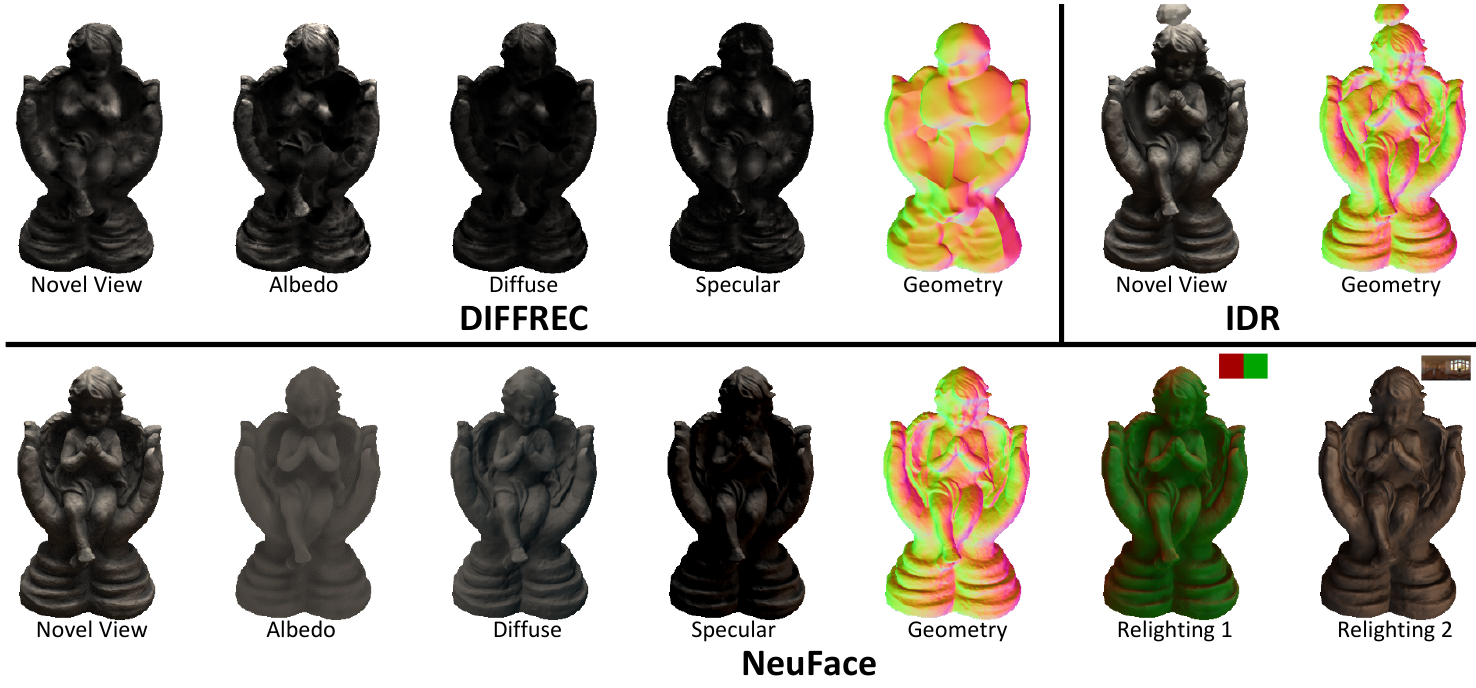}
  \caption{Reconstruction results of scan 118 on the DTU dataset~\cite{jensen2014large} (zoom in for a better view).}
  \label{fig:dtu118}
\end{figure*}

\section{Additional Experiments}

To have a better insight into \textit{NeuFace}, more experimental results are provided.

\subsection{Analysis on Low-Rank Prior}

The low-rank prior is introduced to constrain the solution space. We analyze its effect by varying the number of the bases. As Fig.~\ref{fig:rank} shows, $k=3$ brings the best result. As the number increases, the underlying BRDFs become increasingly intractable to fit with limited observations, leading to performance declines for both metrics. The results validate that the low-rank prior facilitates BRDFs learning. 

\subsection{Miscellaneous Ablation} 
We analyze the neural photometric calibration, the \textit{ImFace} prior, and the specular energy loss. Table~\ref{table:miscellaneous_ablation} demonstrates the importance of each component in achieving realistic face rendering and accurate geometry recovery. Neural calibration remarkably benefits the inverse rendering procedure by accounting for camera inconsistency. The \textit{ImFace} prior significantly improves the recovered geometry and appearance quality by constraining sampling within a valid 3D space during forward rendering. Additionally, the specular energy loss includes real-world empirical evidence on face skin and stabilizes the factorization by preventing energy influx into the specular term.

\subsection{More Results on FaceScape}

\begin{figure*}
  \setlength{\abovecaptionskip}{0pt}
  \setlength{\belowcaptionskip}{0pt}
  \includegraphics[width=1.0\linewidth]{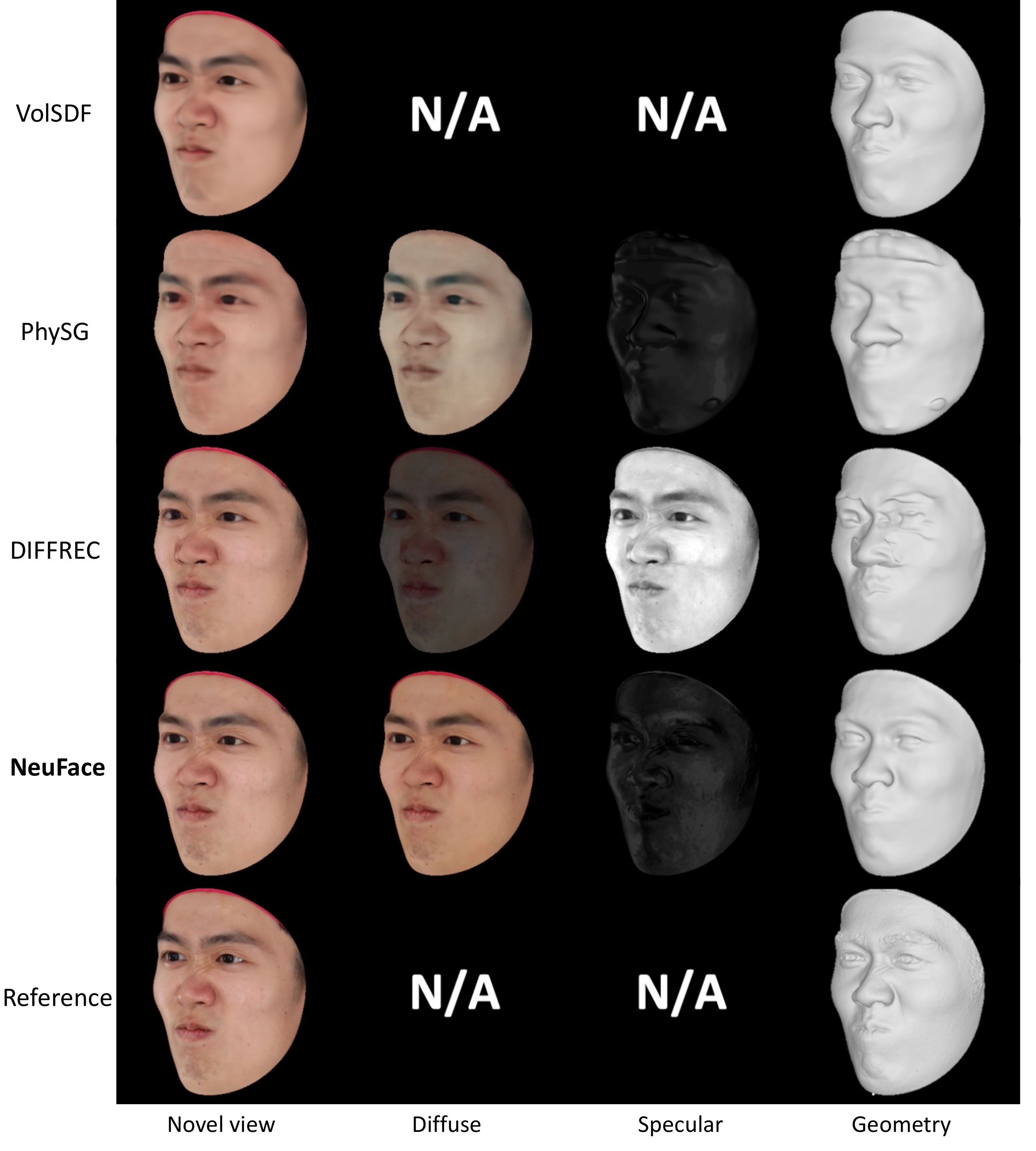}
  \caption{Comparison with VolSDF\cite{yariv2021volsdf}, PhySG\cite{zhang2021physg}, DIFFREC\cite{munkberg2022extracting} on subject 2, expression 4 (anger).}
  \vspace{-5mm}
  \label{fig:212_id_4_exp}
\end{figure*}

\begin{figure*}
  \setlength{\abovecaptionskip}{0pt}
  \setlength{\belowcaptionskip}{0pt}
  \includegraphics[width=1.0\linewidth]{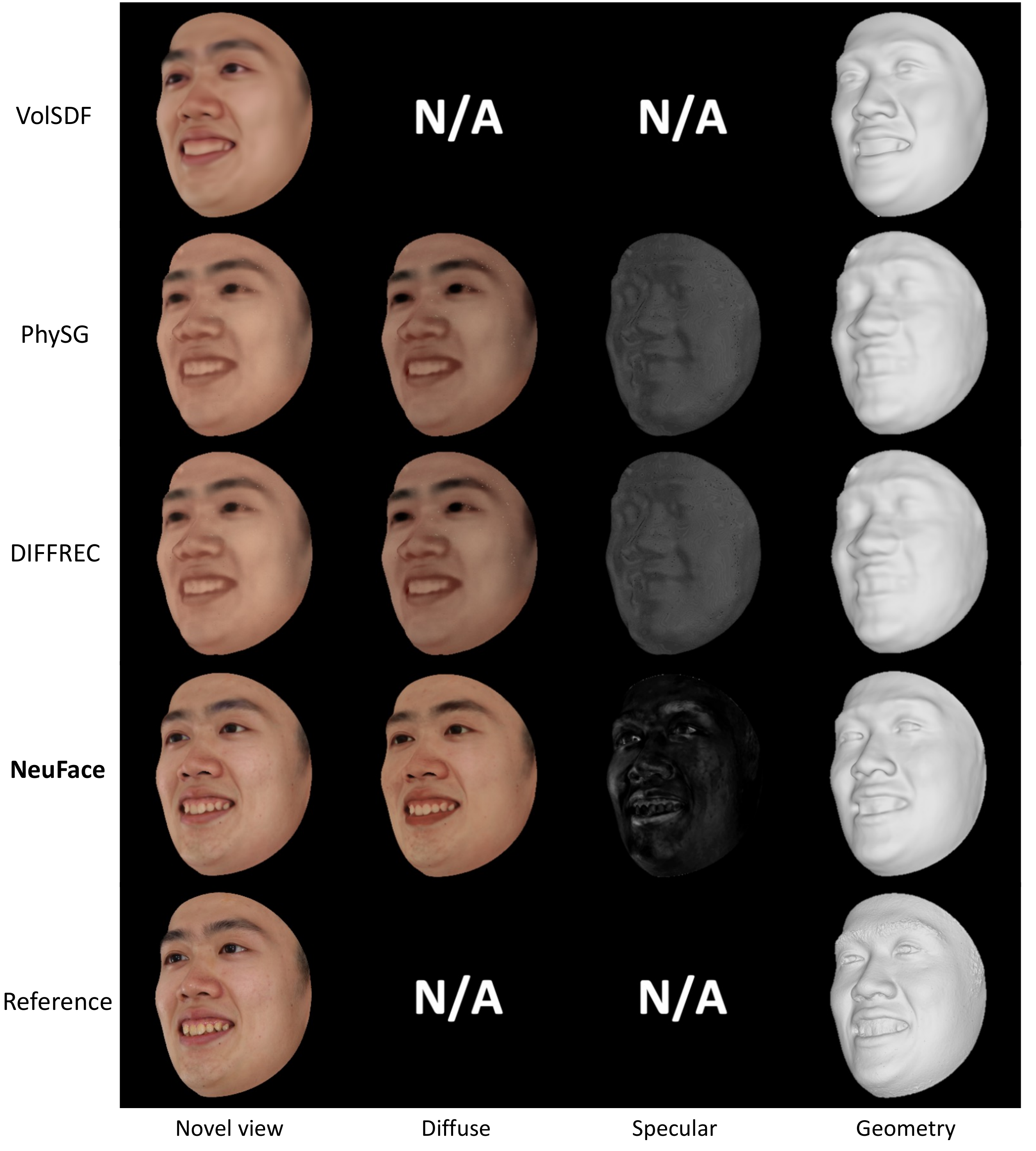}
  \caption{Comparison with VolSDF\cite{yariv2021volsdf}, PhySG\cite{zhang2021physg}, DIFFREC\cite{munkberg2022extracting} on subject 2, expression 16 (grin).}
  \vspace{-5mm}
  \label{fig:212_id_16_exp}
\end{figure*}

\begin{figure*}
  \setlength{\abovecaptionskip}{0pt}
  \setlength{\belowcaptionskip}{0pt}
  \includegraphics[width=1.0\linewidth]{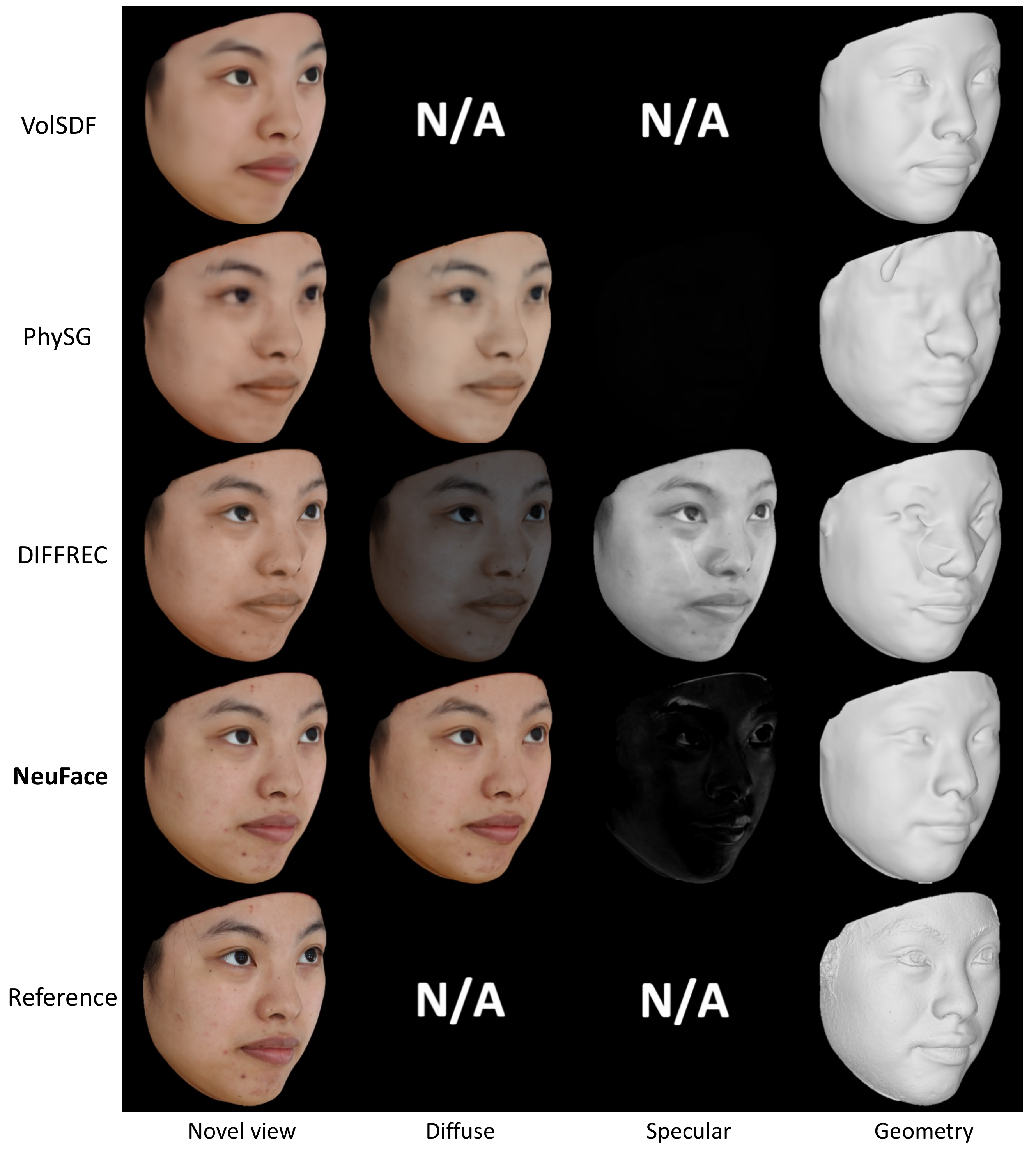}
  \caption{Comparison with VolSDF\cite{yariv2021volsdf}, PhySG\cite{zhang2021physg}, DIFFREC\cite{munkberg2022extracting} on subject 3, expression 1 (neutral).}
  \vspace{-5mm}
  \label{fig:340_id_1_exp}
\end{figure*}

\begin{figure*}
  \setlength{\abovecaptionskip}{0pt}
  \setlength{\belowcaptionskip}{0pt}
  \includegraphics[width=1.0\linewidth]{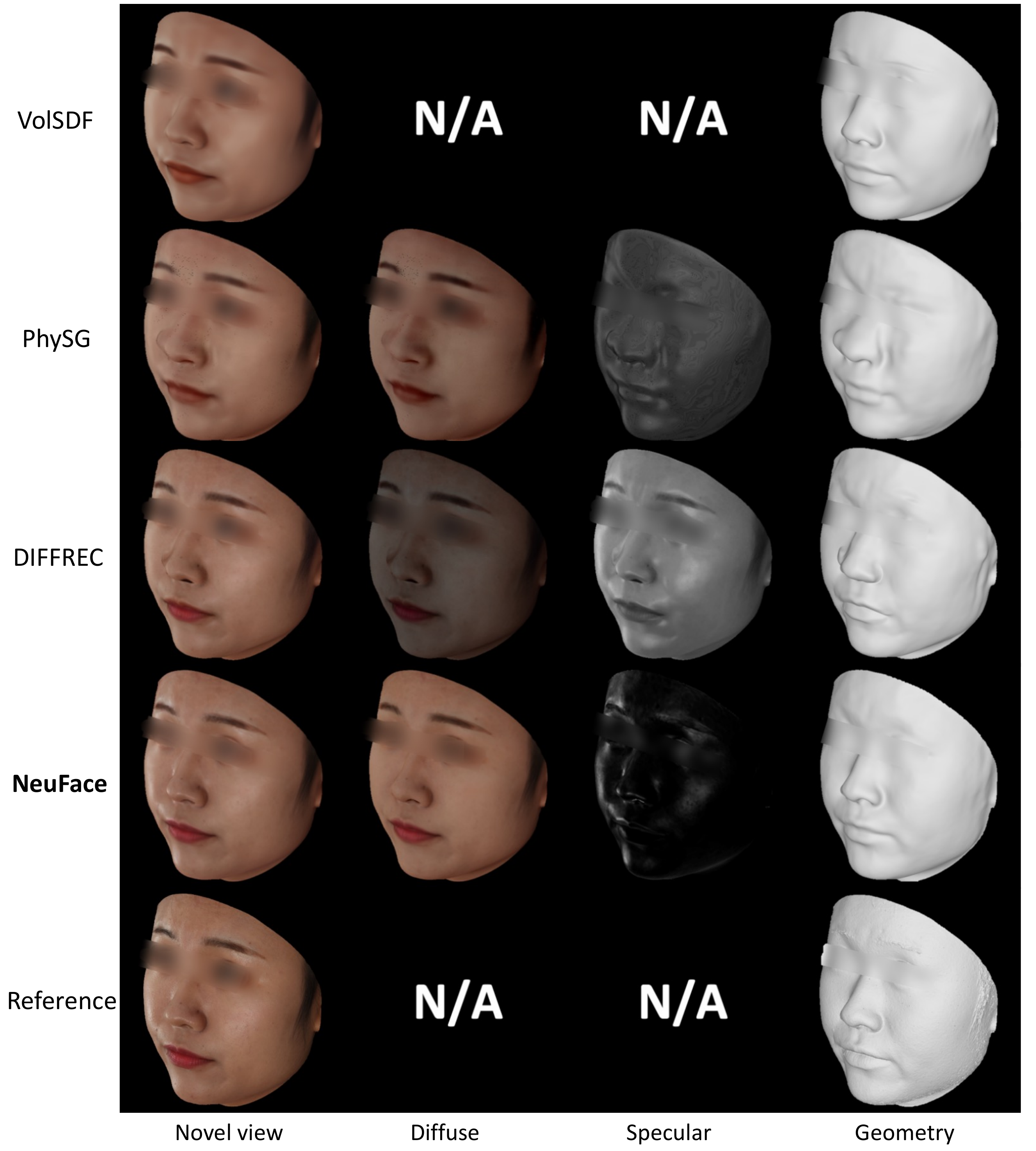}
  \caption{Comparison with VolSDF\cite{yariv2021volsdf}, PhySG\cite{zhang2021physg}, DIFFREC\cite{munkberg2022extracting} on subject 1, expression 2 (smile).}
  \vspace{-5mm}
  \label{fig:1_id_2_exp}
\end{figure*}

In Figs.~\ref{fig:212_id_4_exp}, \ref{fig:212_id_16_exp}, \ref{fig:340_id_1_exp}, and \ref{fig:1_id_2_exp}, we show more comparison with VolSDF\cite{yariv2021volsdf}, PhySG\cite{zhang2021physg}, and DIFFREC\cite{munkberg2022extracting}, where diverse facial expressions are involved (numbered by the FaceScape dataset~\cite{yang2020facescape}). Additionally, Tab.~\ref{tab:supp_result} reports the quantitative metrics evaluated on different expressions. In Tabs.~\ref{tab:ablation_psnr}, \ref{tab:ablation_ssim}, \ref{tab:ablation_lpips}, and \ref{tab:ablation_chamfer}, we also provide the ablation results in terms of PSNR, SSIM, LPILP and the Chamfer distance. For more intuitive perspectives, please refer to the supplementary videos.

\subsection{BRDFs Coefficients Visualization}

To illustrate the underlying distribution of the learned material BRDFs, we visualize the coefficients $\mathbf{c}$ by clustering them into 8 classes with $K$-means. As shown in Fig.~\ref{fig:c_kmeans}, each color indicates a cluster with similar reflections. Visually inspected, \textit{NeuFace} learns various reflection patterns for different areas.

\begin{table}[t]
    \resizebox{1.\linewidth}{!}{
    \begin{tabular}{l|cccccc}
    Shading Model
    & \textbf{S2E1} 
    & \textbf{S2E4}
    & \textbf{S2E16}
    & \textbf{S3E1}
    & \textbf{S3E4}
    & \textbf{S3E16} \\
    \shline
    SH+Phong Model & \cellcolor{lightyellow}{30.28} & \cellcolor{orange}{29.16} & \cellcolor{orange}{28.25} & \cellcolor{orange}{29.99} & \cellcolor{orange}{29.93} & \cellcolor{orange}{29.60}  \\
    Cubemap+Disney BRDF & {29.86} & \cellcolor{lightyellow}{28.67} & \cellcolor{lightyellow}{27.97} & {29.25} & \cellcolor{lightyellow}{29.51} &  \cellcolor{lightyellow}{29.15} \\
    \hline
    Cubemap+Neural Bases  & \cellcolor{orange}{31.15} & {28.36} & {26.89} & \cellcolor{lightyellow}{29.97} & {29.02} & {28.92} \\
    SH+Neural Bases (ours) &  \cellcolor{tablered}\textbf{31.87} & \cellcolor{tablered}\textbf{30.14} & \cellcolor{tablered}\textbf{28.77} & \cellcolor{tablered}\textbf{31.03} & \cellcolor{tablered}\textbf{30.86} & \cellcolor{tablered}\textbf{30.13} \\
    \end{tabular}}
    \vspace{-2mm}
    \caption{Ablation study on shading models in terms of per-scene PSNRs. 
    \vspace{-4mm}
    \label{tab:ablation_psnr}
    }
\end{table}

\begin{table}[t]
    \resizebox{1.\linewidth}{!}{
    \begin{tabular}{l|cccccc}
    Shading Model
    & \textbf{S2E1} 
    & \textbf{S2E4}
    & \textbf{S2E16}
    & \textbf{S3E1}
    & \textbf{S3E4}
    & \textbf{S3E16} \\
    \shline
    SH+Phong Model & \cellcolor{lightyellow}{0.954} & \cellcolor{lightyellow}{0.948} & \cellcolor{orange}{0.946} & \cellcolor{lightyellow}{0.949} & \cellcolor{orange}{0.953} & \cellcolor{lightyellow}{0.945} \\
    Cubemap+Disney BRDF & {0.953} & {0.941} & \cellcolor{lightyellow}{0.938} & {0.945} & {0.951} &  {0.943} \\
    \hline
    Cubemap+Neural Bases  & \cellcolor{orange}{0.959} & \cellcolor{orange}{0.949} & {0.936} & \cellcolor{orange}{0.951} & \cellcolor{lightyellow}{0.952} & \cellcolor{orange}{0.947} \\
    SH+Neural Bases (ours) &  \cellcolor{tablered}\textbf{0.961} & \cellcolor{tablered}\textbf{0.954} & \cellcolor{tablered}\textbf{0.946} & \cellcolor{tablered}\textbf{0.953} & \cellcolor{tablered}\textbf{0.956} & \cellcolor{tablered}\textbf{0.949} \\
    \end{tabular}}
    \vspace{-2mm}
    \caption{Ablation study on shading models in terms of per-scene SSIMs. 
     \vspace{-4mm}
    \label{tab:ablation_ssim}
    }
\end{table}

\begin{table}[t]
    \resizebox{1.\linewidth}{!}{
    \begin{tabular}{l|cccccc}
    Shading Model
    & \textbf{S2E1} 
    & \textbf{S2E4}
    & \textbf{S2E16}
    & \textbf{S3E1}
    & \textbf{S3E4}
    & \textbf{S3E16} \\
    \shline
    SH+Phong Model & {.0409} & \cellcolor{lightyellow}{.0295} & \cellcolor{orange}{.0283} & {.0258} & \cellcolor{lightyellow}{.0265} & \cellcolor{lightyellow}{.0246} \\
    Cubemap+Disney BRDF & \cellcolor{lightyellow}{.0366} & {.0338} & {.0373} & \cellcolor{orange}{.0251} & \cellcolor{tablered}\textbf{.0247} &  \cellcolor{orange}{.0238} \\
    \hline
    Cubemap+Neural Bases  & \cellcolor{orange}{.0320} & \cellcolor{orange}{.0253} & \cellcolor{lightyellow}{.0297} & \cellcolor{lightyellow}{.0256} & \cellcolor{orange}{.0264} & \cellcolor{tablered}\textbf{.0237} \\
    SH+Neural Bases (ours) &  \cellcolor{tablered}\textbf{.0314} & \cellcolor{tablered}\textbf{.0235} & \cellcolor{tablered}\textbf{.0256} & \cellcolor{tablered}\textbf{.0233} & {.0268} & {.0247} \\
    \end{tabular}}
    \vspace{-2mm}
    \caption{Ablation study on shading models in terms of per-scene LPIPSs. 
     \vspace{-4mm}
    \label{tab:ablation_lpips}
    }
\end{table}

\begin{table}[t]
    \resizebox{1.\linewidth}{!}{
    \begin{tabular}{l|cccccc}
    Shading Model
    & \textbf{S2E1} 
    & \textbf{S2E4}
    & \textbf{S2E16}
    & \textbf{S3E1}
    & \textbf{S3E4}
    & \textbf{S3E16} \\
    \shline
    SH+Phong Model & \cellcolor{lightyellow}{0.552} & {0.559} & \cellcolor{lightyellow}{0.491} & \cellcolor{lightyellow}{0.573} & \cellcolor{lightyellow}{0.669} & \cellcolor{lightyellow}{0.588} \\
    Cubemap+Disney BRDF & {0.618} & \cellcolor{lightyellow}{0.530} & {0.614} & {0.704} & {0.764} &  {0.697} \\
    \hline
    Cubemap+Neural Bases  & \cellcolor{tablered}\textbf{0.437} & \cellcolor{tablered}\textbf{0.423} & \cellcolor{tablered}\textbf{0.453} & \cellcolor{tablered}\textbf{0.515} & \cellcolor{orange}{0.596} & \cellcolor{tablered}\textbf{0.515} \\
    SH+Neural Bases (ours) &  \cellcolor{orange}{0.481} & \cellcolor{orange}{0.439} & \cellcolor{orange}{0.467} & \cellcolor{orange}{0.522} & \cellcolor{tablered}\textbf{0.535} & \cellcolor{orange}{0.528}\\
    \end{tabular}}
    \vspace{-2mm}
    \caption{Ablation study on shading models in terms of per-scene Chamfer distances. 
     \vspace{-4mm}
    \label{tab:ablation_chamfer}
    }
\end{table}

\section{Results on DTU}

To evaluate the generalization ability of \textit{NeuFace} to common objects, we perform 3D reconstruction on the DTU MVS dataset~\cite{jensen2014large}. As \textit{ImFace} is not applicable to this case, we slightly modify our pipeline as Fig.~\ref{fig:dtu_pipeline} shows and perform standard surface rendering as in IDR~\cite{yariv2020idr}. For fair comparison, we also apply our neural calibration module to DIFFREC~\cite{munkberg2022extracting}. We implement DIFFREC as depicted in the paper without the tricks in their supplementary material.

As shown in Figs.~\ref{fig:dtu65}, \ref{fig:dtu110} and \ref{fig:dtu118}, \textit{NeuFace} can faithfully recover the appearance and geometry properties of common objects, along with a decent relighting ability. Even though the metals usually have small diffuse energy, satisfactory results are achieved. Moreover, due to the low-rank prior, \textit{NeuFace} is capable of handling the low-textured scan (as Fig.~\ref{fig:dtu110} shows), which is mostly reported as a failure case by existing neural surface reconstruction methods \cite{yariv2020idr, yariv2021volsdf, wang2021neus}.

\section{Ethics Statement}

The proposed approach aims at realistic face capturing, and similar to existing ones, it has the potential to be applied to recover 3D face models from multi-view images, which may lead to a privacy violation problem. We encourage researchers and developers to regulate the illegal spread of 3D face models and consider the questions, such as how to prevent personal face data from being maliciously accessed, before applying the model to the real world. 

It has been confirmed by the FaceScape \cite{yang2020facescape} owner that all the images in this paper can go public. 

\end{document}